\documentclass[lettersize,journal]{IEEEtran}
\usepackage{amsmath,amsfonts}
\usepackage{algorithmic}
\usepackage{algorithm}
\usepackage{array}
\usepackage[caption=false,font=normalsize,labelfont=sf,textfont=sf]{subfig}
\usepackage{textcomp}
\usepackage{stfloats}
\usepackage{url}
\usepackage{verbatim}
\usepackage{graphicx}
\usepackage{cite}

\usepackage{multirow}
\usepackage{amsmath}
\usepackage{amsfonts} 
\usepackage{amssymb}
\usepackage{ragged2e}
\usepackage{booktabs}

\begin{document}

\title{Gaseous Object Detection}

\author{Kailai~Zhou, Yibo~Wang, Tao~Lv,
       Qiu~Shen,~\IEEEmembership{Member,~IEEE,}  Xun~Cao,~\IEEEmembership{Member,~IEEE}

\thanks{Kailai~Zhou, Yibo~Wang, Tao~Lv, Qiu~Shen and Xun~Cao are with Nanjing University, Nanjing, 210023, China.  E-mail: \{calayzhou, ybwang, lvtao\}@smail.nju.edu.cn,  \{shenqiu, caoxun\}@nju.edu.cn}
\thanks{Qiu~Shen and Xun~Cao are the corresponding authors.}
}


\markboth{Journal of \LaTeX\ Class Files,~Vol.~14, No.~8, August~2021}%
{Shell \MakeLowercase{\textit{et al.}}: A Sample Article Using IEEEtran.cls for IEEE Journals}


\maketitle

\begin{abstract}
Object detection, a fundamental and challenging problem in computer vision, has experienced rapid development due to the effectiveness of deep learning. The current objects to be detected are mostly rigid solid substances with apparent and distinct visual characteristics. In this paper, we endeavor on a scarcely explored task named Gaseous Object Detection (GOD), which is undertaken to explore whether the object detection techniques can be extended from solid substances to gaseous substances. Nevertheless, the gas exhibits significantly different visual characteristics: 1) saliency deficiency, 2) arbitrary and ever-changing shapes, 3) lack of distinct boundaries. To facilitate the study on this challenging task, we construct a GOD-Video dataset comprising 600 videos (141,017 frames) that cover various attributes with multiple types of gases. A comprehensive benchmark is established based on this dataset, allowing for a rigorous evaluation of frame-level and video-level detectors. Deduced from the Gaussian dispersion model, the physics-inspired Voxel Shift Field (VSF) is designed to model geometric irregularities and ever-changing shapes in potential 3D space. By integrating VSF into Faster RCNN, the VSF RCNN serves as a simple but strong baseline for gaseous object detection. Our work aims to attract further research into this valuable albeit challenging area.
\end{abstract}


\section{Introduction}

\IEEEPARstart{D}{etecting} gas leaks and emissions holds significant value in numerous fields including climate change \cite{gaalfalk2016making}, atmospheric monitoring \cite{gong2020high} and industry safety \cite{tratt2017mahi} across sectors such as energy, chemicals, and electricity. For instance, reducing greenhouse gas emissions is a major priority in the mission to achieve carbon neutrality by the middle of this century \cite{wigley2011coal}. Moreover, the frequent occurrences of industrial explosions, fires, and air contamination have raised serious societal concerns. Early warning of gas leaks is crucial to avert personnel casualties and economic losses. Recently, the advanced Gas Cloud Imaging (GCI) system has emerged as the industry's next-generation device \cite{hagen2020survey}, which offers superior advantages  such as long range, rapid response, and the unique capability of gas dispersion visualization. The combination of the gas imaging system
 and artificial intelligence  visual analysis shows great promise. Nevertheless, gaseous objects exhibit distinctly different visual characteristics, posing  a  substantial challenge to previous object detection methods \cite{ren2015faster,law2018cornernet,carion2020end}. In this paper, we aim to explore whether the object detection techniques can be extended from solid substances to gaseous substances in the field of computer vision.

\begin{figure}[h]
	\centering
	\includegraphics[width=1.0\linewidth]{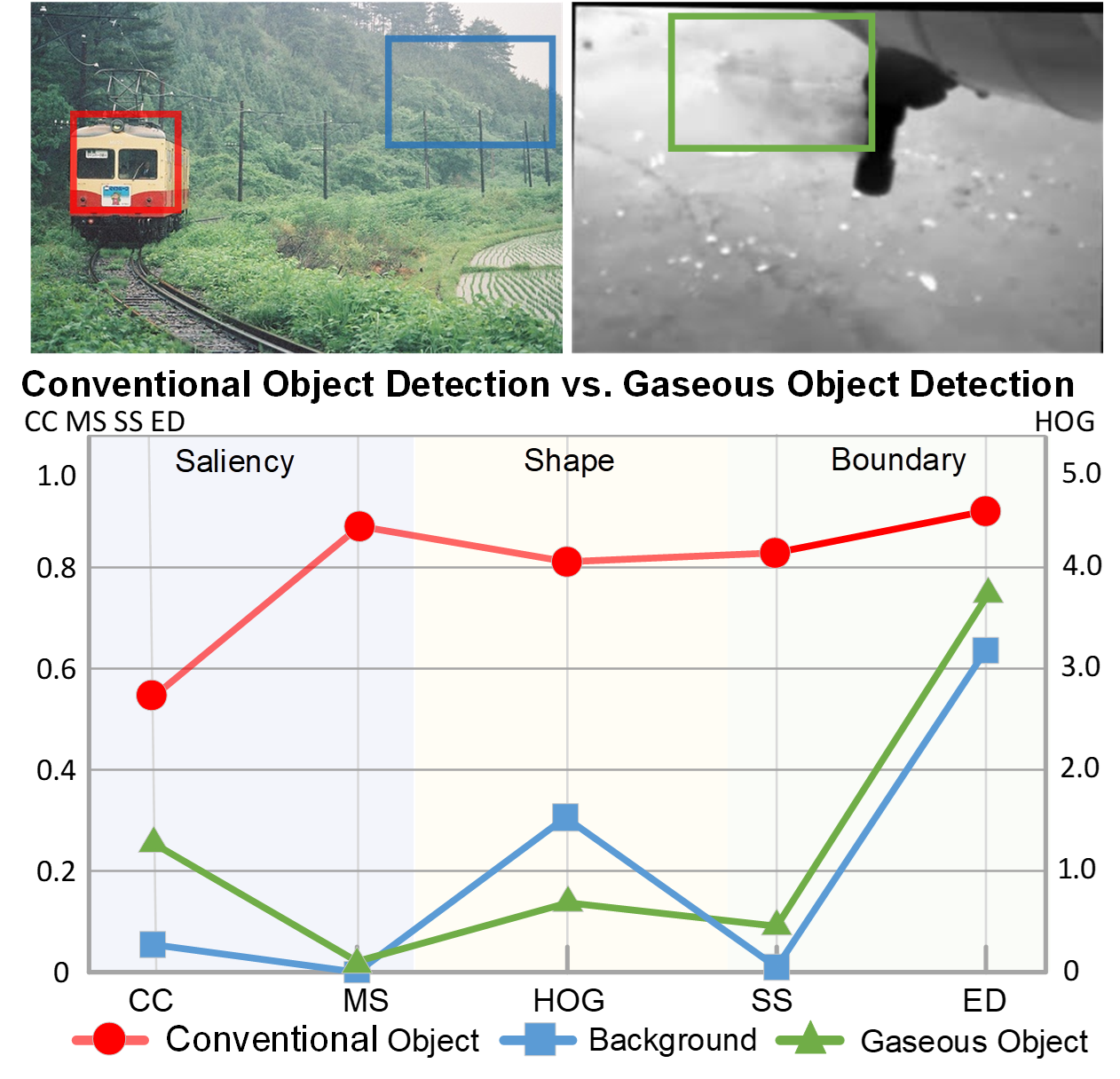}
	\vspace{-2.0em}
	\caption{We compare the conventional object and gaseous object using traditional feature descriptors in three aspects: saliency, shape and boundary. Higher scores of the objectness measures indicate that the box area is more likely to contain an object, while the gaseous object shows greater similarity to the background rather than the foreground.  }
 
	\label{fig:GODFig}
 \vspace{-0.0em}
\end{figure}

Object detection has been a popular research topic that has received continuous attention over several decades \cite{zou2023object}. In response to the inquiry of what an object is, Alexe et al. \cite{alexe2010object} summarize previous research and define a measure of objectness generic over classes, which regards objects as standalone things with a well-defined boundary and center. Based on this observation, four image cues are proposed to distinguish an object: Multi-scale Saliency (MS) \cite{hou2007saliency}, Color Contrast (CC), Edge Density (ED) \cite{canny1986computational}, and Superpixels Straddling (SS) \cite{felzenszwalb2004efficient}. MS and CC  serve as measures to establish  saliency levels, whereas ED and SS help to distinguish the object based on the boundary density and contour information. Additionally, we employ Histograms of Oriented Gradients (HOG) \cite{dalal2005histograms} to elucidate the characteristics of local shapes and structures. As illustrated in Fig.~\ref{fig:GODFig}, the box area with higher scores of above cues is more likely to contain an object. Considering the perspectives of saliency, shape and boundary, we contend that the difference between gaseous objects and conventional objects lies in the following aspects:


1) Saliency deficiency: The gaseous object exhibits low saliency in multi-scale maps, appearing more similar to the background with respect to the MS and CC measurements.

2) Arbitrary and ever-changing shapes: The pattern of gaseous objects is influenced by various factors such as leakage rate, wind speed, and the surrounding environment. These factors give gaseous objects the liberty to assume arbitrary shapes that change continuously over time.

3) Lack of distinct boundaries: The characterization of gaseous objects as closed boundaries is challenging due to the weak edge magnitudes and lack of distinct contours.

Unlike the conventional object with distinct visual appearances, traditional objectness measures suggest that the gaseous object exhibits greater similarity to the background rather than the foreground. The limited visual cues within a single frame image pose great difficulties for existing object detection methods \cite{ren2015faster,law2018cornernet,carion2020end}. Considering that the dynamic gas diffusion is more pronounced at the video level, traditional visual-based methods for gas detection primarily employ background modeling \cite{zeng2018gas,wang2018infrared} and optical flow  \cite{hagen2020survey}. These methods require that cameras remain stationary and  that gas concentrations be relatively high, which are inadequate for addressing complex and variable gas leaks in practical industrial environments. Given the difficulty in explicitly modeling gaseous characteristics with handcrafted descriptors, a plausible solution involves implicitly extracting spatio-temporal features with deep video-level detectors. Nevertheless, existing video-level detectors \cite{jiao2021new} primarily  rely on  single key frame information for region proposals, which may be insufficient  for localization in the GOD task. Extracting collaborative spatio-temporal representation in the region proposal stage  to accurately model the geometric irregularities and ever-changing shapes of gaseous objects presents a significant challenge.

In addition, since  gas leaks are rare events that are extremely difficult to collect, there is a notable scarcity of large-scale, high-quality datasets. To facilitate the study of this challenging task, we have constructed a  pioneering dataset named GOD-Video, specifically designed for gaseous object detection. This dataset comprises 600 videos with a total of 141,017 frames, covering a broad range of distances, sizes, visibility levels, and spectral ranges for different types of gases. Furthermore, we conduct a comprehensive statistical comparison  between the objects in the GOD-Video and COCO \cite{lin2014microsoft} datasets in terms of saliency, shape and boundary. We hope our dataset will be a valuable resource for the computer vision community.



Based on the GOD-Video dataset, we have developed a fair and comprehensive benchmark, which aims to provide deep insights into the performance of  existing frame-level and video-level detectors on the GOD task. Inspired by the physical properties of the gas diffusion process, we introduce the voxel shift field, which predicts a 3D offset $(dx, dy, dt)$ that enables the arbitrary shift along the x, y, and t axes for each voxel in the feature map. By means of learnable and flexible shift strategy in the potential 3D space, the VSF facilitates the adaptive modeling of saliency deficiency and geometric irregularities in the spatial dimension, as well as ever-changing and arbitrary shapes in the temporal dimension. The VSF can be seamlessly integrated into the majority of established object detectors, imparting conventional 2D detectors with 3D modeling capabilities. With the combination of VSF and Faster RCNN \cite{ren2015faster}, the VSF RCNN increases the $AP_{50}$ from 37.34\% to 51.08\%, demonstrating a substantial improvement over the Faster RCNN baseline. The contributions of this paper are summarized as follows:


$\bullet$ We create the pioneering GOD-Video dataset comprising 600 videos (141,017 frames), which provides a large-scale, high-quality, and diversified foundation for gaseous object detection.  Considering the perspectives of saliency, shape, and boundary, we analyze the unique characteristics of gaseous objects by statistical comparisons between the GOD-Video and COCO datasets.

$\bullet$ We conduct a comprehensive evaluation of both frame-level and video-level detectors, and the spatio-temporal behaviors of action recognition methods in the GOD task are investigated in a unified framework.

$\bullet$ Deduced from the Gaussian dispersion model, the physics-inspired voxel shift field is specifically designed to capture geometric irregularities and ever-changing shapes in the potential 3D space, demonstrating its adaptability across various detectors. The VSF RCNN serves as a simple but strong baseline and exhibits the capability to model temporal irregular shapes in relevant tasks.

\section{RELATED WORK}

\subsection{Object Detection }
Object detection has been a fundamental and challenging
problem in computer vision. Faster RCNN \cite{ren2015faster} is the cornerstone detector in the deep learning era that generates and refines region proposals in a unified learning framework. To overcome the computation redundancy of two-stage detectors \cite{cai2018cascade,ren2015faster}, one-stage detectors \cite{redmon2018yolov3,dai2016r,lin2017focal} aim to strike a balance between accuracy and speed. Anchor-free detectors are proposed to avoid manual tuning of anchor configurations and can be classified into two types: anchor-point detectors \cite{tian2019fcos,zhang2020bridging} and key-point detectors \cite{zhou2019objects,dong2020centripetalnet}. Recently, DETR-based detectors \cite{carion2020end, zhu2020deformable, zhang2022dino} have emerged as a new paradigm,  eliminating the need for many hand-engineered components through the transformer's self-attention mechanism. To effectively leverage the temporal information, we survey two relevant areas: video object detection and spatio-temporal action detection.

Video Object Detection. Feature degradation, such as motion blur, occlusion, and defocus, presents the primary challenge of video object detection. Early box-level video object detection methods tackle this problem in a post-processing way by linking bounding boxes predicted by still frames \cite{kang2017t, han2016seq, feichtenhofer2017detect, chen2018optimizing}. Feature-level video object detection methods aggregate temporal contexts to improve the feature representation. The feature can either be improved at the image level to boost the single-frame detector \cite{bertasius2018object, xiao2018video, zhu2017flow, zhu2018towards,zhu2017deep} or at the object level through exploration of semantic and spatio-temporal correspondence among the region proposals \cite{wu2019sequence, deng2019relation, shvets2019leveraging, yao2020video, han2020mining, deng2019object, gong2021temporal,wang2018fully}.  To avoid the redundant computational cost of applying object detectors to every frame, previous video object detection methods have focused on propagating useful information from key frame features to non-key frame features. 

Spatio-temporal Action Detection. The action detection task aims to identify and localize human actions in videos. Two-stage action detectors \cite{tang2020asynchronous,wu2020context,pan2021actor} first predict the bounding boxes of actors and then perform actor-centric action recognition, such as Context-Aware RCNN \cite{wu2020context}.
The end-to-end action detectors \cite{sun2018actor,girdhar2019video,li2020actions,zhao2022tuber,chen2021watch} simultaneously train the actor proposal network and the action classification network. For example, MOC \cite{li2020actions} jointly optimizes localization and classification losses based on the video feature maps which concatenates frame features along the temporal axis. STMixer \cite{wu2023stmixer} dynamically integrates video features across both spatial and temporal dimensions, with its cross-attention decoder independently processing spatial and temporal queries. Action detection methods utilize 2D-CNN or classic 3D-CNN architectures, such as I3D \cite{carreira2017quo} and SlowFast \cite{feichtenhofer2019slowfast}, as the feature extraction backbone, and focus on temporal interaction based on the extracted features in a post-process manner.

In conclusion, previous video-level detectors primarily rely on established frameworks for proposing region proposals. They pay more attention to the temporal relationship modeling after localization to achieve a deeper understanding of video content. However, this paradigm might be unsatisfactory for the GOD task, which necessitates considering the unique gaseous characteristics and requires more emphasis on collaborative representation of spatio-temporal features in the region proposal stage. 

\subsection{Spatio-temporal Feature Extraction}

Various techniques for action recognition \cite{chen2021deep}, such as two-stream networks, 3D-CNNs, and compute-effective 2D-CNNs have been explored for the extraction of spatio-temporal information. The classic paradigm for two-stream networks incorporates extra modalities including optical flow, which acts as a second input pathway to capture temporal motion details \cite{cheron2015p,wang2016temporal}. 3D-CNNs facilitate the direct extraction of spatio-temporal information from unprocessed input streams, explicitly representing spatio-temporal features\cite{tran2015learning,carreira2017quo,hara2018can}. Nevertheless, the high computational demands of 3D convolution kernels have led researchers to explore 3D factorization techniques to reduce complexity \cite{qiu2017learning, tran2018closer}. Alternatively, frame-level features can be extracted using 2D-CNNs, followed by modeling of temporal correlations. Compute-effective 2D-CNNs employ operations such as temporal shift \cite{fan2019more,shao2020temporal,lin2019tsm}, low-cost difference operations \cite{wang2021tdn} and establishing correspondences across adjacent frames \cite{kwon2020motionsqueeze}. While these methods are primarily tailored for video classification tasks, there remains a lack of exploration of spatio-temporal architecture suitable for video-level detection tasks.

\subsection{Gas Leak Detection}
Traditional gas leak detection primarily employs point sensors \cite{murvay2012survey}, similar to electronic noses, which require the gas to diffuse into the sensor for identification and have a limited range ($\leq$ 10 m). In contrast,  gas imaging cameras \cite{hagen2020survey,gaalfalk2016making} serve as ``intelligent eyes", which offer distinct advantages such as extensive monitoring coverage, rapid response speeds, and the unique capability of gas dispersion visualization. The potential of gas imaging technology can be substantially enhanced  with advanced artificial intelligence visual analysis. Early attempts for visual-based gas detection involve background modeling \cite{zeng2018gas,wang2018infrared,hong2019vocs} and optical flow methods \cite{hagen2020survey}. Background modeling  captures the dynamic changes of gases across a sequence of frames, but it suffers from the interference of moving objects and requires that the cameras remain stationary. Optical flow estimates how each parcel of gas moves within the image based on the ``brightness constancy" assumption \cite{fleet2006optical}, which necessitates the gas to have a relatively high concentration. Traditional hand-crafted descriptors struggle to handle the complex and variable characteristics of gaseous objects.

A plausible solution involves implicitly extracting gaseous features using deep-learning based methods. VideoGasNet \cite{wang2022videogasnet} considers it a video classification problem and classifies the videos by methane leak volume. TBLD \cite{bin2021tensor} first takes advantage of the tensor decomposition based background subtraction algorithm to identify the foreground area, and then investigates different classifiers in the leakage classification stage. The above works focus on utilizing deep learning techniques for video gas classification tasks.  Additionally, some works \cite{bin2022multimodal,bin2022foreground,shi2020real} adopt the classic object detector, such as Faster RCNN \cite{ren2015faster}, for localizing gaseous objects. However, the frame-level detector cannot leverage the temporal correlation for the dynamic diffusion of gases, underscoring the need for the spatio-temporal feature extraction specifically designed for gaseous object detection. In conclusion,  research in this area is still in its nascent stages and remains relatively scarce.

\begin{figure}[t]
	\begin{center}
		\includegraphics[width=1.0\linewidth]{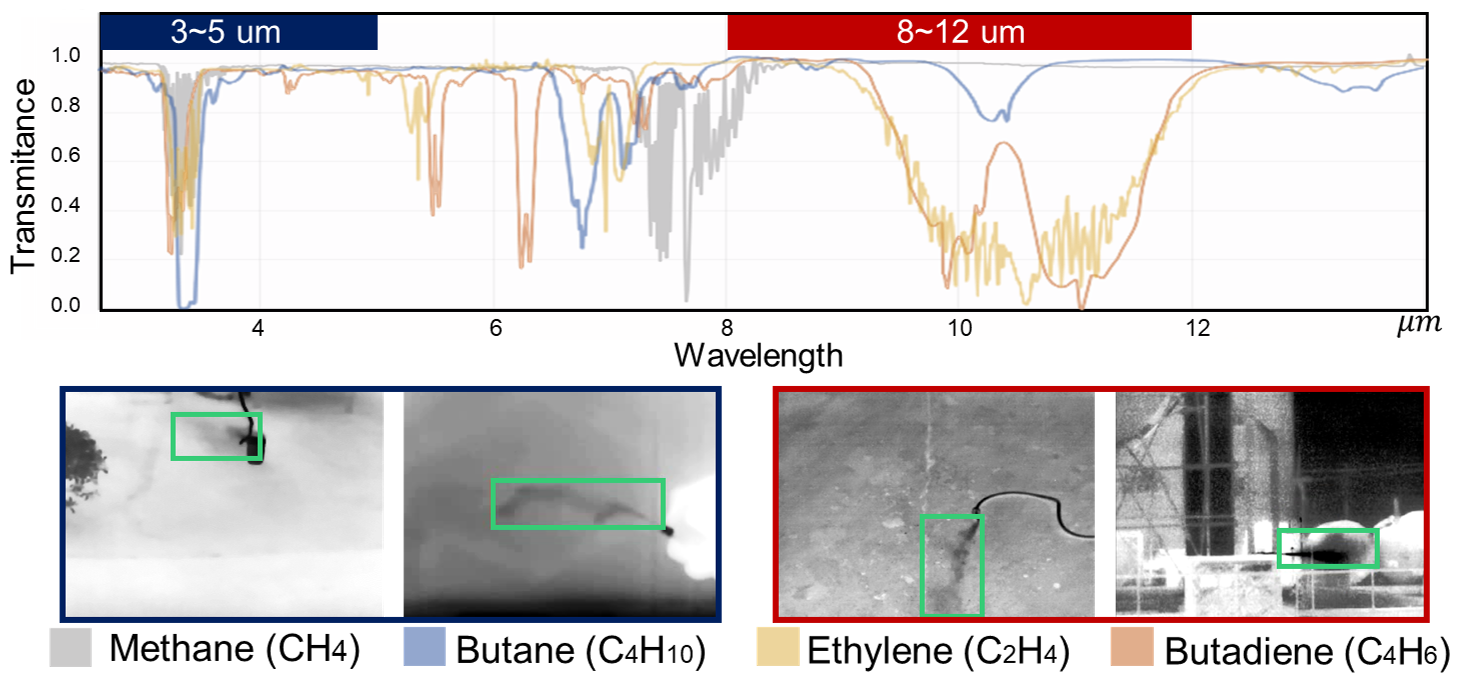}
	\end{center}
	\vspace{-1.0em}
	\caption{The spectral transmittance curves of representative gases in the mid-infrared band. Despite differences in gas types and spectral ranges, based on the Lambert-Beer's law, they exhibit similar visual characteristics in gas imaging cameras.  }
	\label{fig:principle}
		\vspace{-0.0em}
\end{figure}

\section{GOD-VIDEO DATASET}
To facilitate the study on the GOD task, this section details the collection, properties, and comparative analysis of the GOD-Video dataset.

\subsection{Gas Imaging Principle}
Many gases are invisible to human eyes but exhibit unique spectral absorption characteristics in the mid-infrared band, often referred to as the spectral fingerprint region. As light passes through the gas cloud, the vibration of polyatomic molecules induces changes in the dipole moment, resulting in the absorption of infrared spectra. The gas molecules absorb light energy at their characteristic frequencies, a relationship that follows Lambert-Beer's law.  As depicted in Fig.~\ref{fig:principle}, regions containing gaseous objects appear darker than the surrounding background when narrowband filtering is applied in the 3-5 $\mu m$ or 8-12 $\mu m$. Under the same conditions, the more pronounced the characteristic absorption peaks of gases within a fixed wavelength range, the more distinct the difference between the gas regions and the background in gas imaging cameras. Our GOD-Video samples are captured using either a 3-5 $\mu m$ or 8-12 $\mu m$ gas imaging camera.

\begin{figure*}[t]
	\begin{center}
		\includegraphics[width=1.0\linewidth]{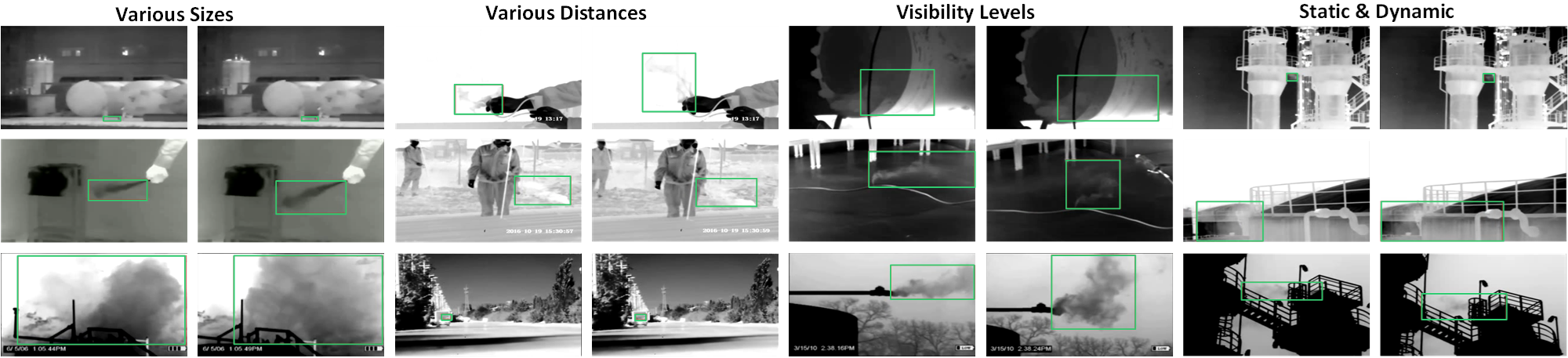}
	\end{center}
	\vspace{-1.0em}
	\caption{We show the representative samples of different attributes in the GOD-Video dataset. The green rectangles represent the annotated boxes.  }
	\label{fig:dataset_overview}
		\vspace{-0.0em}
\end{figure*}
\begin{figure*}[t]
	\begin{center}
		\includegraphics[width=0.95\linewidth]{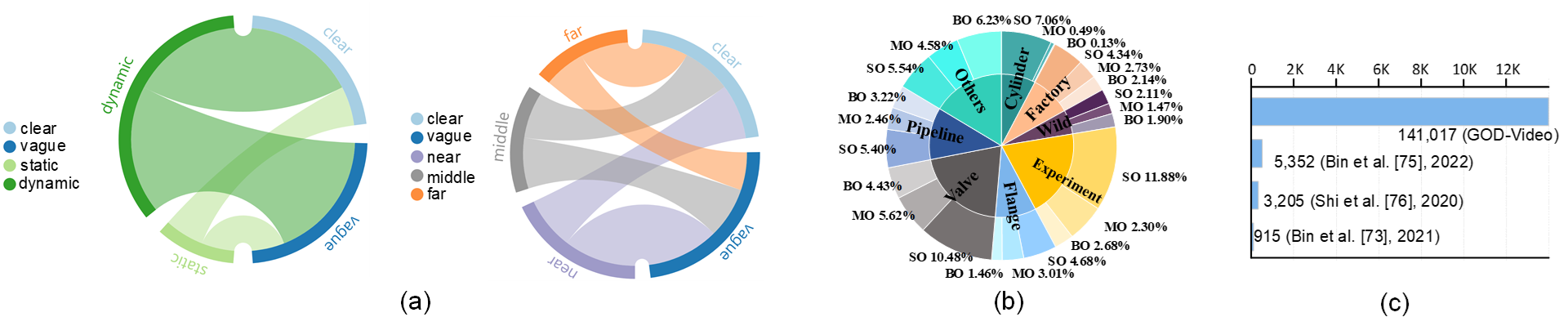}
	\end{center}
	\vspace{-1.0em}
	\caption{GOD-Video dataset details. (a) Mutual dependencies among attributes. (b) Scene taxonomy of the GOD-Video dataset (SO: Small Object, MO: Middle Object, BO: Big Object). (c) Comparisons with previous gas leak detection datasets in the frame-level. The quantity of samples in the GOD-Video dataset surpasses previous datasets significantly.}
	\label{fig:dataset}
		\vspace{-0.0em}
\end{figure*}

\subsection{Data Collection and Annotation}
Data Collection. Samples in the GOD-Video dataset are collected in two ways: 1) Manual inspection, where inspection personnel carry handheld portable gas imaging cameras for identifying gas leaks at industrial sites; 2) Deployed devices, where gimbal-mounted cameras installed at chemical plants enable continuous, 24/7 real-time monitoring of hazardous gases. Given that gas leaks are low-probability  events, it takes over three years to accumulate our dataset,  which demands significant human and material resources.

Data Preprocessing. Firstly, we trim the clips by timeline to ensure that each frame contains gaseous objects. Secondly, we divide the varying lengths of video clips into shorter segments of approximately 10 seconds. Thirdly, we resize all the videos with different spatial resolutions to $320\times 240$ to maintain a uniform image size. The data preprocessing procedure is manually curated as follows: 1) A maximum of two representative clips from each scene at different times are selected to ensure dataset diversity. 2) Clips with poor image quality or significant  noise are discarded. 3) Clips that are visually indistinguishable are removed. The final GOD-Video dataset consists of 600 video samples, totaling 141,017 frames.

Data Annotation. For the intuitive and practical purposes, rectangular bounding annotations are employed for the GOD task. We implement a two-phase process that aims to annotate the gaseous object as accurately as possible.  In the first phase, GOD-Video is meticulously annotated by three experienced annotators using a specially developed tool. This tool enhances annotator efficiency and accuracy by providing pseudo-color, motion information from a background modeling algorithm \cite{stauffer1999adaptive}, and historical frames spanning two seconds. The extra information significantly improves the annotators' ability to identify gaseous objects. In the second phase, frame-level annotations are double-checked for consistency according to established rules: 1) Annotations are temporally continuous without sudden change.  2) Bounding boxes tighten the object boundary well according to human's subjective perception. 3) Bounding boxes reacts immediately when diffusion direction varies. Since the displacement of the bounding boxes between adjacent frames is minimal, the samples are labeled every five frames, and the annotation of intermediate frames is obtained through linear interpolation. The representative samples and corresponding bounding boxes are illustrated in Fig.~\ref{fig:dataset_overview}.

\subsection{Dataset Properties}
In terms of the captured objects, environments, equipment and practical applications, the GOD-Video dataset considers a variety of attributes to ensure diversity.

Gas Types: Gaseous objects in the GOD-Video primarily include  alkanes ($\mathrm{C_nH_{2n+2}}$) and alkenes ($\mathrm{C_nH_{2n}}$), which are fundamental to the petrochemical industry.   Fig.~\ref{fig:principle} shows these gases  adhere to Lambert-Beer’s law and exhibit similar visual characteristics. Therefore, they are uniformly classified  as gaseous objects in this paper. It is worth noting that this category-independent definition is based on the object morphological features in computer vision, extending beyond the concept of gases in the physical world.

Spectral Ranges: The 3-5 $\mu m$ and 8-12 $\mu m$ bands are chosen based on atmospheric windows and the characteristic absorption peaks of typical gases. The 3-5 $\mu m$ band corresponds to the characteristic absorption of alkane gases, while the 8-12 $\mu m$ corresponds to that of alkene gases.

Shooting Scenes: The GOD-Video dataset comprises eight distinct scenes: pipeline, factory, flange, valve, experiment, cylinder, wild, and others. Notably, the cylinder and experiment scenes correspond to samples of gas release from cylinders and human-induced gas emissions, respectively. The remaining scenes are categorized according to different industrial backgrounds.

Object Sizes: The GOD-Video dataset covers objects of different sizes, classified into small (area $\textless$ 32×32), medium ( 32×32 $\textless$ area $\textless$ 96×96), and large (area $\textgreater$ 96×96) according to the COCO prototype \cite{lin2014microsoft}. These different-sized objects are scattered throughout various scenes, displaying a good uniformity in Fig.~\ref{fig:dataset} (b).

Different Distances: Fig.~\ref{fig:dataset_overview} shows GOD-Video spans shooting distances from a few meters to several hundred meters.

Visibility Levels: The clarity of gas imaging is influenced by multiple factors, including gas types, leakage rates, capturing  devices, distances, and backgrounds. For example, under the same conditions, butane appears more distinct than methane due to its stronger absorption rate in the characteristic band.  GOD-Video samples are divided into clear and vague subsets according to whether annotators can judge the  object boundary within a single frame.

Camera Motion States:  The GOD-Video dataset comprises two camera states, static and dynamic, captured with handheld and deployed devices to more effectively meet practical requirements.

Fig.~\ref{fig:dataset} (a) demonstrates multi-dependencies among GOD-Video attributes, the larger width of a link between two super-classes indicates a higher probability. For example, the dynamic vague samples occupy a larger proportion than the clear static samples. GOD-Video dataset achieves good diversity by providing various distances, sizes, visibility levels and scenes captured by different spectral ranges.

\begin{figure*}[t]
	\begin{center}
		\includegraphics[width=1.0\linewidth]{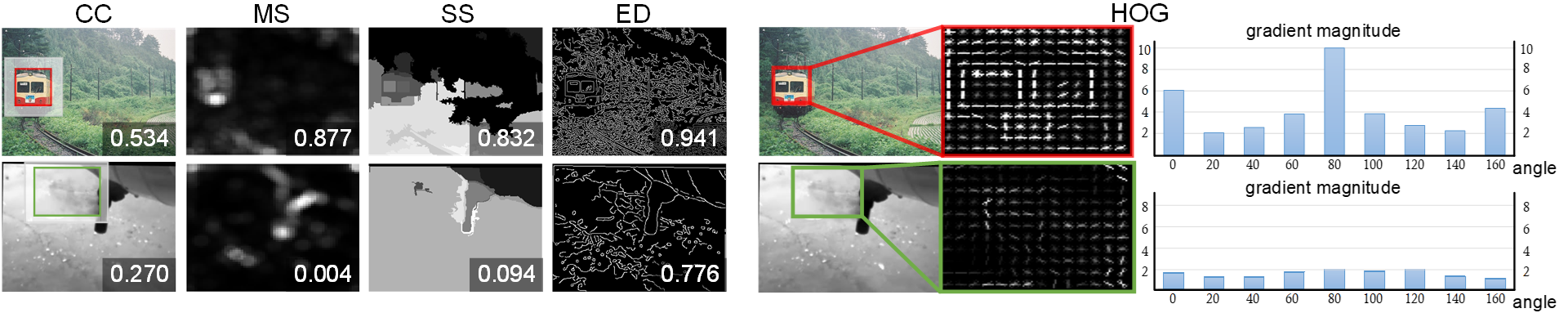}
	\end{center}
	\vspace{-1.5em}
	\caption{ The visualization of objectness measures (CC, MS, SS, ED) with corresponding scores, and  the visualization of HOG for the train and gas samples. }
	\label{fig:cue}
		\vspace{-0.0em}
\end{figure*}

\subsection{Comparisons with Existing Datasets}

In Fig.~\ref{fig:dataset} (c), we present a comparison between GOD-Video and previous gas detection datasets \cite{bin2021tensor,bin2022foreground,shi2020real}. Research on gaseous objects remains in its early stages, with existing datasets primarily limited to frame-level samples and a small total number of samples.  For example, the dataset by Bin et al. (2021) \cite{bin2021tensor} comprises only 915 images that stem from merely three 30-second video segments, whereas the dataset by Shi et al. (2020) \cite{shi2020real} contains 3205 images from 11 video segments. In contrast, the samples in the GOD-Video dataset are at the video-level, allowing for the complete exploitation of temporal information to enhance detection accuracy. Additionally, previous datasets only cover a narrow range of gas types. For instance, Bin et al.’s (2021) dataset \cite{bin2021tensor} focuses exclusively on leaked natural gas, while Shi et al.’s (2020) dataset \cite{shi2020real} is limited to ethane leaks at industrial sites. The GOD-Video dataset not only is two orders of magnitude larger than previous datasets but also offers significantly greater diversity, establishing a robust foundation for the gaseous object detection.

\subsection{Gaseous Object vs. Conventional Object}

To attain a deeper understanding  of gas characteristics, we conduct a statistical analysis between  gaseous objects in the GOD-Video dataset and conventional objects in the COCO dataset \cite{lin2014microsoft}. It is considered that any object has at least one of three distinctive characteristics \cite{alexe2010object}: 1) a well-defined closed boundary in space; 2) a different appearance from their surroundings; 3) sometimes it is unique within the image and stands out as salient. Consequently, our evaluation with the traditional descriptors is based on three perspectives: saliency, boundary and shape. Initially, we present scores and visualizations of objectness measures for the train and gas samples in Fig.~\ref{fig:cue}:

1) Saliency. Multiscale Saliency (MS) \cite{hou2007saliency} indicates that an object should be a salient region with a unique appearance. Color Contrast (CC) reflects the color dissimilarity between the foreground and the background. Typically, conventional objects generally exhibit conspicuous saliency traits. For instance, the train achieves a score of 0.877 with the MS measure in Fig.~\ref{fig:cue} (a). However, the gaseous object lack saliency in the multi-scale map, which only achieves a score of 0.004.


2) Boundary. Edge Density (ED) calculates the average edge magnitude as closed boundary characteristics, as they tend to have many edgels in the inner ring \cite{alexe2010object}.  The Superpixels Straddling (SS) cue employs superpixels  \cite{felzenszwalb2004efficient}  as features to divide the image into small regions of uniform color or texture, and it yields a score of 0.094 when applied to the gaseous object. The lack of distinct borders presents a difficulty to characterize the gaseous object accurately using the SS cue. Similarly, when the gas concentration is low, the ED cue tends to reflect the edge information of the background rather than that of the gas. Different from the objects in COCO dataset, the relatively sparse edges and poorly defined contours of gaseous objects pose a challenge. 

3) Shape. Histograms of Oriented Gradients (HOG) \cite{dalal2005histograms} capture edge or gradient structures, which are the characteristics of local shapes. The visualization in Fig.~\ref{fig:cue} indicate that the gradient distribution of the train is primarily concentrated in the horizontal and vertical directions. In contrast, gas gradients are scattered across various angles and appears more chaotic. Additionally, the gradient magnitude of gas is significantly weaker than the train. While the shape of conventional solid objects remains relatively stable, gaseous objects exhibit patterns influenced by factors such as leakage rate, wind speed, and surrounding environment. These factors grant gases the ability to assume arbitrary shapes that continuously evolve over time.

\begin{figure}[t]
	\begin{center}
		\includegraphics[width=1.0\linewidth]{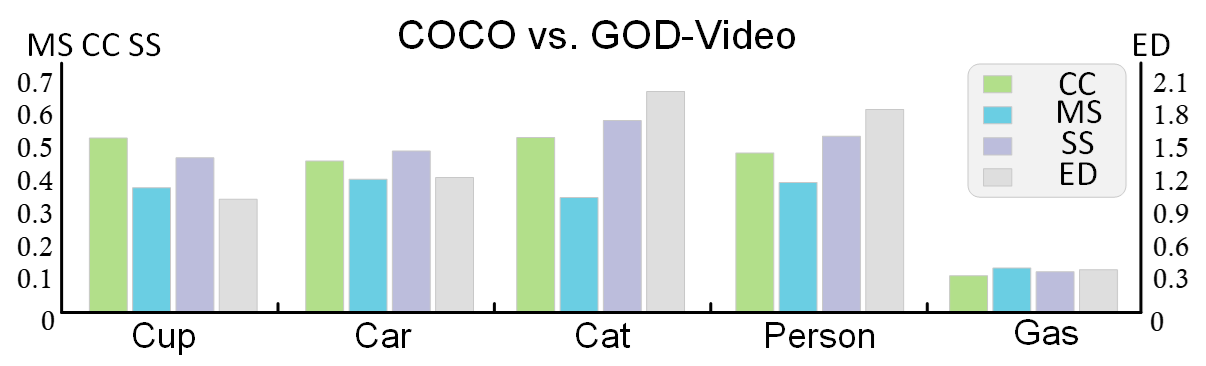}
	\end{center}
	\vspace{-1.0em}
	\caption{ A statistical analysis of objectness measures is conducted between the  typical classes (cup, car, cat, person) in the COCO \cite{lin2014microsoft} dataset and the gas in the GOD-Video dataset.  }
	\label{fig:coco}
		\vspace{-0.0em}
\end{figure}

Then we analyze the unique characteristics of gaseous objects by performing statistical comparisons between the GOD-Video and COCO \cite{lin2014microsoft} dataset. We select 1,000 objects in the given category from the COCO and GOD-Video dataset, and calculate the mean score to establish the metric of evaluation in a statistical manner. From the statistical result in Fig.~\ref{fig:coco}, the average score of gas is significantly lower than that of categories in the COCO dataset.
Given that gaseous objects exhibit limited image cues in a single frame, it is crucial to consider the dynamic nature of gas diffusion when modeling features in the spatio-temporal domain.

\section{METHOD}
To address the challenges of saliency deficiency, arbitrary and ever-changing shapes, and the lack of clear boundaries, we first analyze the physical characteristics of gas diffusion with the Gaussian dispersion model \cite{holmes2006review,bosanquet1936spread}. Subsequently, we present technical details of the voxel shift field, which is derived from the Gaussian dispersion model. Lastly, we integrate the voxel shift field into Faster RCNN \cite{ren2015faster} to create VSF RCNN, which grants conventional 2D detectors the ability to model features in the potential 3D space.

\begin{figure*}[t]
	\begin{center}
		\includegraphics[width=0.95\linewidth]{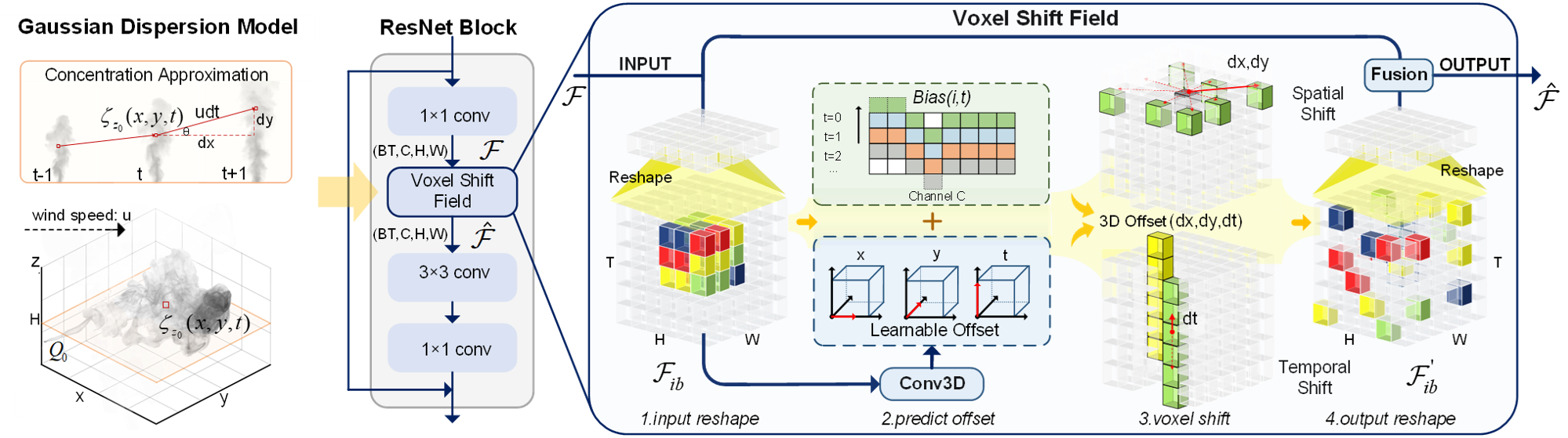}
	\end{center}
	\vspace{-1.0em}
	\caption{ We deduce from the widely adopted Gaussian dispersion model \cite{bosanquet1936spread} that under certain conditions, the gas concentration $\zeta_{z_0}{(x, y, t)}$  at the given spatial location $(x, y)$ and time $t$ can be appropriately estimated by the concentration $\zeta_{z_0}{(x + dx, y + dy, t + dt)}$ in an adjacent location $(x+dx, y+dy)$ at time $t+dt$. Building upon this principle, the physics-inspired voxel shift field first predicts a 3D offset composed of a learnable offset and an initial bias along the temporal axis, then performs the spatio-temporal shift operation to each voxel in the feature map $\mathcal{F}_{ib}$,  enabling the modeling of geometric irregularities and ever-changing shapes of gaseous objects. 
 }
	\label{fig:vsf}
		\vspace{-0.0em}
\end{figure*}

\subsection{Gaussian Dispersion Model}
Upon release from a source, gas diffusion exemplifies fundamental fluid motion characteristics. Fick's law of diffusion indicates that the rate of change of concentration at a particular location with respect to time is proportional to the second derivative of concentration with respect to distance. The Gaussian dispersion model \cite{bosanquet1936spread} can be deduced by establishing boundary conditions \cite{stockie2011mathematics}, and is presently one of the most established and widely-used gas dispersion models, due to its simple mathematical expressions and direct physical significance. The Gaussian dispersion model is divided into the Gaussian plume model and the Gaussian puff model \cite{slade1968meteorology}. The Gaussian puff model is appropriate for depicting the diffusion of lightweight gases from continuous sources or in instances where the release time is equal to or greater than the diffusive timescale. By modeling the plume as a series of puffs, the Gaussian puff model, extensively applied in dynamic gas simulation, is expressed as follows:

\begin{equation}
\begin{aligned}
\begin{split}
&\zeta{(x, y, z, t)}=\frac{Q_0}{2 \pi^{\frac{3}{2}} \sigma_x \sigma_y \sigma_z} 
e^{\frac{-(x \cdot \cos \theta+y \cdot \sin \theta-u t)^2}{2\sigma_x^2}} \cdot\\
&e^{\frac{-(y \cdot \cos \theta-x \cdot \sin \theta)^2}{2\sigma_y^2}}\left[e^{\frac{-(z+H)^2}{2\sigma_z^2}}+e^{\frac{-(z-H)^2}{2\sigma_z^2}}\right]
\end{split}
\end{aligned}
\end{equation}


\noindent where $ \zeta{(x,y,z,t)}$ represents the predicted gas concentration at any downstream point $ (x, y, z) $ at time $t$. $Q_{0}$ represents the source strength, $u$ represents the ambient wind speed, $\theta$ is the angle of wind direction with the x axis, $H$ represents the effective leak height, and $\sigma_x$, $\sigma_y$, $\sigma_z$ are the gas dispersion coefficients in the x, y and z directions, given by the Pasquill-Gifford dispersion coefficient equation \cite{pasquill1961estimation}. 
In the vicinity of the instantaneous time $dt$, we can assume that the gas diffusion coefficients $\sigma_x$, $\sigma_y $ and $\sigma_z$ are constant. Therefore, for the cross section at height $z=z_{0}$, we set  $\mathcal{Q}=\frac{Q_0}{2 \pi^{\frac{3}{2}} \sigma_x \sigma_y \sigma_z} $ and  $\mathcal{Z}=e^{\frac{-(z_0+H)^2}{2\sigma_z^2}}+e^{\frac{-(z_0-H)^2}{2\sigma_z^2}}$, the $\zeta{(x, y, z_0, t)}$ can be rewritten as $\zeta_{z_0}{(x, y, t)}$:


\begin{equation}
\begin{aligned}
\begin{split}
\zeta_{z_0}{(x, y, t)}= \mathcal{Q} \cdot
e^{\frac{-(x \cdot \cos \theta+y \cdot \sin \theta-u t)^2}{2\sigma_x^2}} \cdot
e^{\frac{-(y \cdot \cos \theta-x \cdot \sin \theta)^2}{2\sigma_y^2}} \cdot \mathcal{Z}
\end{split}
\end{aligned}
\end{equation}

\noindent where the $\mathcal{Q}$  and $\mathcal{Z}$ can be considered as the constant when $dt$ is approaching 0. Due to the 
continuity of gas diffusion in the spatio-temporal domain, we aim to approximate the gas concentration $\zeta_{z_0}{(x, y, t)}$ with the neighboring concentration $\zeta_{z_0}{(x+dx, y+dy, t+dt)}$ when $dx$, $dy$ and $dt$ satisfy certain conditions:


\begin{equation}
\left\{\begin{array}{c}
\mathrm{d} x \cdot \cos \theta+\mathrm{d} y \cdot \sin \theta - \mathrm{u} \cdot \mathrm{d} t  = 0\\
\mathrm{~d} y \cdot \cos \theta-\mathrm{d} x \cdot \sin \theta   = 0 
\end{array}\right.
\label{eqn:3}
\end{equation}

 It can be inferred from Eq. (\ref{eqn:3}) that the $dx, dy$ is determined by the wind speed $u$ and angle $\theta$. The gas concentration $\zeta_{z_0}{(x, y, t)}$ at the spatial position $(x, y)$ at time $t$ can be approximately expressed by the gas concentration $\zeta_{z_0}{(x+dx, y+dy, t+dt)}$ at the neighboring spatial position $(x+dx, y+dy)$ at time $t+dt$:

\begin{equation}
\zeta_{z_0}{(x, y, t)}=\zeta_{z_0}{(x + dx, y + dy, t + dt)}
\label{eqn:4}
\end{equation}

The Eq. (\ref{eqn:4}) demonstrates that although  the concentration of gas may be feeble in a particular frame, it can be enhanced by neighboring spatial positions in adjacent frames. For practical applications, one factor that needs to be considered is the influence of camera motion states.  As the rigid body motion, the camera movement can be transformed from the world coordinate to the camera coordinate via rotation matrices and translation vectors. In the camera coordinate, over very small time interval, the motion of the camera can be integrated into the gas dispersion process. Thus, the collective offset of dynamic camera motion and gas dispersion can be modeled as a whole. Consequently, we propose the voxel shift field base on the Gaussian dispersion model.



\subsection{Voxel Shift Field}
Fig.~\ref{fig:vsf} illustrates the overall architecture of the voxel shift field. Given that GOD is a video-level detection task, the feature map of multiple frames spans four dimensions, represented by $\mathcal{F} \in \mathbb{R}^{BT\times C\times H\times W}$. Here, $B$ is the batch size,  $C$ denotes the number of channels, $H$ and $W$ are the spatial dimensions, and $T$ represents the number of input frames. The input of voxel shift field is the feature map after $1\times 1$ convolution in the ResNet block \cite{he2016deep}, since the channel dimension is reduced and the computational cost will be lower. For the input feature map ${\mathcal{F}}\in \mathbb{R}^{BT\times C\times H\times W }$, we first reshape its dimension to ${\mathcal{F}}\in \mathbb{R}^{B\times C\times H\times W \times T }$. We denote the feature map of the i-th channel and b-th batch as ${\mathcal{F}_{ib}} \in \mathbb{R}^{H\times W \times T}$. Then the voxel shift field predicts the 3D offset for each voxel in the feature map ${\mathcal{F}_{ib}}$. The 3D offset parameters for ${\mathcal{F}_{ib}}$ along the x, y and t  axes are expressed as $ \mathcal{O} \in \mathbb{R}^{H \times W \times T \times 3}$, which  consist of two parts: the shift values predicted by the 3D convolution applied to the feature map 
$\mathcal{F}$, and the initial temporal bias $Bias(i,t)$, which simulates   shift operations in action recognition network \cite{lin2019tsm}:

\begin{equation}
\mathcal{O} =Conv3D(\mathcal{F}) + Bias(i, t)
\end{equation}


With the learned 3D offset $(dx, dy, dt)$ for the voxel at coordinate $(x, y, t)$, we approximate and aggregate this voxel according to the concentration approximation formula (\ref{eqn:4}) of the Gaussian dispersion model. Specifically, the voxel in the shifted feature map ${{\mathcal{F}^\prime}_{ib}}{(x, y, t)}$ is derived as follows:



\begin{equation}
\begin{split}
{\mathcal{F}_{ib}}^\prime{\left(x, y, t\right)}= {\mathcal{F}_{ib}}{\left(x+dx, y+dy, t+dt\right)}
\end{split}
\end{equation}

\noindent where the voxel at the coordinate $(x, y, t)$ in the shifted feature map ${{\mathcal{F}_{ib}}^\prime}$ is approximated by the voxel at the adjacent coordinate $(x+dx, y+dy, t+dt)$ in the input feature map $\mathcal{F}_{ib}$. 
We implement the above shifting operation for each voxel in the ${\mathcal{F}_{ib}}$. This design will 
 facilitate the modeling of geometric irregularities and ever-changing shapes of gaseous objects. Another issue that can arise is that, because the predicted offset values in the x, y and t directions are floating-point data types, the sampling position may not align with a voxel perfectly. Hence, it is necessary to obtain $\mathcal{F}_{ib}{(x+dx,y+dy,t+dt)}$ through bilinear interpolation, as shown in Eq. (\ref{eqn:9}):

\begin{small}
\begin{equation}
\begin{split}
&{\mathcal{F}_{ib}}{\left( x+\mathrm{d} x, y+\mathrm{d} y, t+\mathrm{d} t\right)}= \\
&\sum_{n=1}^N \mathrm{~w}(\Delta x, \Delta y, \Delta t) {\mathcal{F}_{ib}}{\left(x+d x+\Delta x, y+d y+\Delta y, t+d t+\Delta t\right)}
\end{split}
\label{eqn:9}
\end{equation}
\end{small}

\noindent in this context, $N=8$ denotes the eight corner points of the grid cube surrounding the sampling point.  $(\Delta x, \Delta y,\Delta t)$ represent the distances from the n-th corner point to the sampling point. The weight coefficient $w(\Delta x,\Delta y,\Delta t)$ used in bilinear interpolation is determined by the distances of $\Delta x$, $\Delta y$ and $\Delta t$. We implement the voxel shift field for each voxel across all batches and channels of the input feature map $\mathcal{F}$. To maintain the consistent size with the input feature map ${\mathcal{F}}$, the shifted feature map ${\mathcal{F}}^\prime \in \mathbb{R}^{B\times C\times H\times W\times T}$ is reshaped back to the   ${\mathcal{F}}^\prime  \in \mathbb{R}^{BT\times C\times H\times W}$.  Furthermore, considering that sampling areas beyond the boundary can result in feature values becoming zero, the output feature map $\hat{\mathcal{F}}$ of VSF is the fusion of  the shifted feature map $\mathcal{F}^{\prime}$ and the input feature map $\mathcal{F}$ with the differential attention:


\begin{equation}
\hat{\mathcal{F}}=\mathcal{F}^{\prime} \oplus\left(\sigma\left(\operatorname{FC}\left(\operatorname{GAP}\left(\mathcal{F}^{\prime}-\mathcal{F}\right)\right)\right) \otimes \mathcal{F}\right)
\end{equation}
\noindent where $\operatorname{GAP}$ represents global average pooling, $\operatorname{FC}$ represents fully connected layer, $\sigma$ is the activation function, $\otimes$ represents element-wise multiplication, and $\oplus$ represents element-wise addition. The voxel shift field increases the 3D receptive field by  enhancing information interaction across consecutive frames. Due to the highly flexible and learnable displacement strategy, the voxel shift field enables the modeling of geometric irregularities and ever-changing shapes while efficiently exploiting the spatio-temporal representation in the potential 3D space.

\subsection{VSF RCNN}
\begin{figure}[t]
	\begin{center}
		\includegraphics[width=1.0\linewidth]{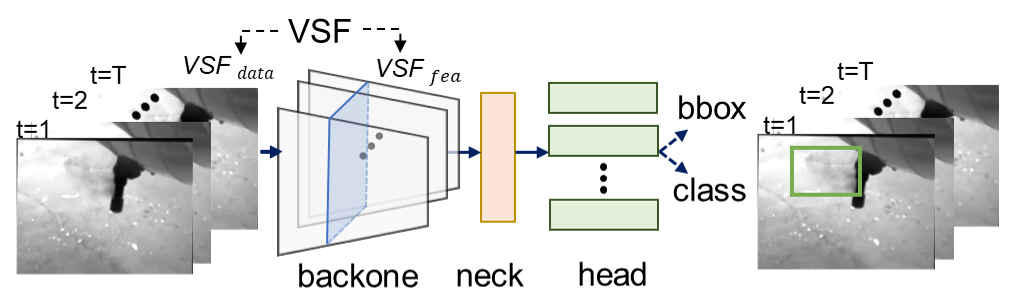}
	\end{center}
	\vspace{-1.0em}
	\caption{ The  overall architecture of VSF RCNN for gaseous object detection. }
	\label{fig:vsfrcnn}
		\vspace{-0.0em}
\end{figure}

As illustrated in Fig.~\ref{fig:vsfrcnn}, most existing object detectors consist of three components: backbone, neck, and head. Unlike conventional object detection, gaseous object detection necessitates the extraction of collaborative spatio-temporal representations from multiple frames for accurate localization. To adapt existing object detectors to the GOD task with minimal modifications, we collaboratively extract multi-frame features at the backbone level, and the detection head responsible for classification and regression tasks is independently constructed for each individual frame.

The VSF RCNN is developed with the integration of voxel shift field into Faster RCNN \cite{ren2015faster}. The voxel shift field serves two primary functions. Firstly, it performs temporal shift operations on the input data to promote information interaction along the t-axis, denoted as $VSF_{data}$. Secondly, it executes spatio-temporal shift operations for the feature map of each ResNet block \cite{he2016deep} in the potential 3D space, referred as $VSF_{fea}$. To maintain consistency with RGB image inputs, we replicate the single-channel infrared image into three channels. $VSF_{data}$ achieves a more effective data representation by temporal misalignment across three channels.
The 3D offset $\mathcal{O}$ in $VSF_{data}$ consists only of an initial temporal shift $Bias(i, t)_{data}$, while $Conv3D(\mathcal{F})$ is set to zero. The $Bias(i, t)_{data}$ is defined as follows:

\begin{equation}
Bias(i, t)_{data}=\left\{\begin{array}{cc}
-1, &  i=0,\ 1 \leq t < T \\
T-1, & i=0,\ t = 0 \\
0, &  i=1,\ 0 \leq t < T  \\
+1, & i=2,\ 0 \leq t < T-1 \\
1-T, & i=2,\ t = T-1
\end{array}\right.
\end{equation}



In each ResNet block, the channel dimension of the feature map is reduced  through  the $1\times 1$ convolution. Subsequently, the $VSF_{fea}$ is employed to manipulate the voxel in the 3D space. The voxel shift field serves as an additional embedded module that does  not change the size of the input feature map. The 3D offset $\mathcal{O}$ in $VSF_{fea}$ consists of  $Conv3D(\mathcal{F})$ and  $Bias\left(i,t\right)_{fea}$. The initial $Bias\left(i,t\right)_{fea}$ in $VSF_{fea}$ acts only on the temporal dimension,  with different initial offset values assigned to various channels.  We set the maximum  initial offset value at 2 to broaden the temporal receptive range. The $Bias(i, t)_{fea}$ is defined as follows:

\begin{equation}
Bias(i, t)_{fea}=\left\{\begin{array}{cc}
-2, & 0 \leq i< \frac{1}{8}C \\
\vspace{0.2em}
-1, & \frac{1}{8}C \leq i< \frac{1}{4}C\\
\vspace{0.2em}
+1, & \frac{1}{4}C \leq i< \frac{3}{8}C \\
\vspace{0.2em}
+2, & \frac{3}{8}C \leq i<\frac{1}{2}C \\
\vspace{0.2em}
0, & \frac{1}{2}C \leq i< C
\end{array}\right.
\end{equation}


In the final output of the backbone, each feature map undergoes channel-dimension reduction using a 1 $\times$ 1 convolution, and is then concatenated to form a multi-frame feature representation for the Region Proposal Network (RPN). During the RPN stage, the optimization target is the mean coordinate value across multiple frames. The loss function during the RPN stage is represented as follows:

\begin{equation}
\mathrm{L}(\{p\},\{g\})=L_{c l s}\left(p, p^*\right)+\lambda p^* L_{r e g}\left(g_t, \frac{1}{T} \sum_1^T g_t^*\right)
\end{equation}

\noindent where $p$ denotes the probability of the anchor box predicted as a positive sample, $p^{\ast}$ represents the ground truth of the anchor box. Additionally, $g_{t}$ indicates four parameterized coordinates of the predicted bounding box in the t-th frame, and $g^{\ast}_{t}$ denotes  the ground truth coordinates of the positive anchor box. The hyper-parameter $\lambda$ represents the balancing parameter between the classification and regression losses, which is set to 1 by default.
The RPN stage generates initial region proposals to be further refined in the detection head. Given the total of T frames in the input, T detection heads are created for the VSF RCNN. The t-th detection head utilizes the corresponding feature from the t-th frame as input and predicts the final box coordinates based on the initial region proposals from the RPN stage. The VSF RCNN is optimized through the two-stage loss of RPN and T detection heads as a whole.

\section{EXPERIMENTS}

\begin{table*}[ht!]
\caption{Evaluation of the previous frame-level detectors and video-level detectors based on the GOD-Video dataset. The GOD detectors are the methods specifically designed for the gaseous object detection task. All detectors are implemented with the same evaluation metric.}
\vspace{-1.5em}
\begin{center}
\renewcommand{\arraystretch}{1.2}
\setlength{\tabcolsep}{1.9mm}{
\begin{tabular}{c|c|c|cccccccc}
\hline
                                                                                        & Detector               & Backbone       & $AP_{50}$  & $AP_{75}$  & $AP_{clear}$ & $AP_{vague}$ & $AP_{s}$ &$AP_{m}$& $AP_{l}$& $AP$    \\ \hline
\multirow{15}{*}{\begin{tabular}[c]{@{}c@{}}Frame-level \\      Detectors\end{tabular}}  & Faster RCNN\cite{ren2015faster}   & ResNet50\&FPN  & 36.15 & 6.82  & 17.41     & 9.60      & 3.88  & 8.28  & 22.27 & 13.08 \\
                                                                                        & Cascade RCNN \cite{cai2018cascade}           & Resnet50\&FPN  & 36.18 & 10.16 & 20.48     & 9.91      & 5.63  & 9.85  & 24.23 & 14.72 \\
                                                                                        & R-FCN \cite{dai2016r}                    & ResNet50\&FPN  & 28.24 & 3.68  & 12.46     & 6.49      & 2.47  & 5.29  & 16.67 & 9.19  \\
                                                                                        & YOLOv3 \cite{redmon2018yolov3}                 & DarkNet53      & 32.25 & 5.06  & 13.69     & 7.68      & 4.53  & 6.79  & 19.12 & 12.31 \\
                                                                                        & YOLOv7 \cite{wang2023yolov7}                 & CSPDarknet53
      &  33.25 &  8.37  &  16.72 & 10.54  &  4.63  &  9.01 &  22.12 & 13.26  \\

                                                                                        & SSD \cite{liu2016ssd}                    & VGG16\&FPN \cite{simonyan2014very}     & 34.72 & 6.41  & 16.21     & 8.98      & 3.58  & 7.66  & 20.87 & 12.23 \\
                                                                                        & RetinaNet \cite{lin2017focal}              & ResNet50\&FPN  & 37.13 & 7.23  & 18.70     & 9.56      & 3.72  & 8.67  & 23.35 & 13.51 \\
                                                                                        & FCOS \cite{tian2019fcos}                   & ResNet50\&FPN  & 34.78 & 6.80  & 17.25     & 8.86      & 2.91  & 8.25  & 22.49 & 12.64 \\
                                                                                        & ATSS \cite{zhang2020bridging}                   & ResNet50\&FPN  & 37.44 & 9.14  & 20.25     & 10.15     & 4.78  & 9.94  & 24.71 & 14.64 \\
                                                                                        & CenterNet \cite{zhou2019objects}             & DLA34 \cite{yu2018deep}          & 32.48 & 7.14  & 16.36   & 8.48  &    4.56 & 7.44  &  22.42&12.04\\
                                                                                        & Centripetalnet \cite{dong2020centripetalnet}        & HourglassNet \cite{newell2016stacked} &  34.03  & 8.58 & 18.33
                                                                             &9.18 &5.07  &9.03  &22.50 &13.28    \\ 

                                                                                        & Sparse RCNN \cite{sun2023sparse}        & ResNet50\&FPN &  31.85 
  & 4.73   & 15.80   & 7.00   & 4.25   & 8.24   & 19.18 & 10.67     \\                                                                                 
                                                                                             & DETR \cite{carion2020end}        & ResNet50 &  26.41 
  & 3.75  & 12.90   & 5.62  & 2.36  & 6.70  & 16.65  & 8.66    \\     
                                                                                          & Deformable DETR 
 \cite{zhu2020deformable}        & ResNet50 &  31.11    & 5.15 
  & 16.35      & 6.55  & 4.08  & 7.77   & 20.14    & 10.75      \\     
                                                                                          & DINO
 \cite{zhang2022dino}        & ResNet50 &  31.95   & 6.34 
  & 17.14     & 7.63    & 3.81  & 8.43  & 20.76   & 11.63     \\     
                                                                       
                                                                             \hline
\multirow{8}{*}{\begin{tabular}[c]{@{}c@{}}Video-level   \\      Detectors\end{tabular}} & DFF \cite{zhu2017deep}      & ResNet50\&FPN  &  32.62 &	5.71 &	14.25 &	8.93 &	0.41 &	6.26 &	21.60 & 	11.38    \\
                                                                                        & FGFA \cite{zhu2017flow}                   & ResNet50\&FPN  &    35.26 &	6.53& 	16.57 &	9.03 &	0.91 &	7.33 &	23.05 &	12.50      \\
                                                                                        & SELSA \cite{wu2019sequence}                  & ResNet50\&FPN  &     36.80& 	7.60 &	18.64 &	9.49 &	1.30 &	8.77 	&24.43 &	13.67     \\
                                                                                        & TRA \cite{gong2021temporal}                    & ResNet50\&FPN  &   36.06 &	7.76& 	18.53 &	9.51 &	0.88 &	8.42 &	24.95 &	13.57    \\
                                                                                        
                                                                                        
                                                                                        & Context-aware RCNN \cite{wu2020context}              & ResNet101\&FPN & 36.15 & 7.46  & 18.84     & 9.14      & 3.92     & 9.14     & 22.39     & 13.52 \\
                                                                                        & MOC \cite{li2020actions}                    & DLA34 \cite{yu2018deep}          & 36.81 & 8.94  & 19.96     & 9.62      & -     & -     & -     & 14.29 \\
                                                                                        & MOC+Flow \cite{li2020actions}              & DLA34 \cite{yu2018deep}          & 34.50 & 7.28  & 18.18     & 8.49      & -     & -     & -     & 12.75
                                                                                        \\ 

                                                                                        & STMixer \cite{wu2023stmixer}              & SlowFast-R50 \cite{feichtenhofer2019slowfast}          & 32.60 
 &4.02   &16.67      & 6.10      & 2.72     & 8.09   & 18.64   & 10.35 
                                                                                        \\

                                                                                        \cline{1-11}
\multirow{3}{*}{GOD Detectors}                                                                    & CenterNet (TEA) \cite{zhou2022explore}   & Res2Net50 \cite{gao2019res2net}      & 42.19 & 8.66  & 22.97     & 9.95      & -     & -     & -     & 15.69 \\
                                                                                        & CenterNet (TEA+STAloss) \cite{zhou2022explore}& Res2Net50 \cite{gao2019res2net}      & 45.08 & 9.50  & 24.43     & 10.91     & -     & -     & -     & 16.99 \\ 
                                                                                        & VSF RCNN (ours)             & ResNet50\&FPN  & \textbf{51.08} & \textbf{12.97 }& \textbf{28.99}     & \textbf{13.02 }    & \textbf{6.67 } & \textbf{15.17} & \textbf{31.56} & \textbf{20.43} \\ \hline 
\end{tabular}}
\label{tab:benchmark}
\end{center}
\vspace{-0.0em}
\end{table*}

\subsection{Experimental Settings}
\label{sec:ES}
We develop our VSF RCNN based on MMDetection \cite{chen2019mmdetection}, a widely recognized open-source object detection toolbox. The data augmentation adapted from MOC \cite{li2020actions} is implemented at the video clip level, including operations such as mirroring, distorting, expanding, and cropping. During training, we crop a clip patch with the size of [0.3, 1]  and resize it to 288 $\times$ 288. Subsequently, each clip is randomly distorted and horizontally flipped with a probability of 0.5 to increase diversity. The entire network is trained using the SGD optimizer, with a learning rate of 2e-2 and a batch size of 16 on two NVIDIA 3090 GPUs, each with 24GB memory. We decrease the learning rate by 0.1x at the 8th epoch, and training concludes at the 9th epoch. For video-level detectors, unless otherwise specified, we set the number of input frames to 8. 

To ensure a fair comparison between frame-level and video-level detectors, we adhere to the COCO protocol \cite{lin2014microsoft} and utilize Average Precision ($AP$) as the evaluation metric. We consider object sizes ($AP_{s}$, $AP_{m}$, $AP_{l}$) and visibility levels ($AP_{vague}$, $AP_{clear}$) as critical factors in gaseous object detection. These factors are influenced by variables such as gas type, concentration, wind speed, leakage rate, and background conditions. Accordingly, we define the following eight metrics: $AP_{50}$ ($AP$ at IoU = 0.5), $AP_{75}$ ($AP$ at IoU = 0.75), $AP_{clear}$ ($AP$ for the clear set), $AP_{vague}$ ($AP$ for the vague set), $AP_{s}$ ($AP$ for small objects, area $\textless$ 32×32), $AP_{m}$ ($AP$ for middle objects, 32×32 $\textless$ area $\textless$ 96×96), $AP_{l}$ ($AP$ for large objects, area $\textgreater$ 96×96), and overall $AP$ ($AP$ over all IoU thresholds). Furthermore, we randomly partition the GOD-Video dataset into three splits, maintaining a train-test ratio of approximately 2:1. We employ K-Fold cross-validation to report the results, averaging them across these three splits in accordance with the common setting \cite{li2020actions,kalogeiton2017action} on the J-HMDB dataset \cite{jhuang2013towards}.

\subsection{Benchmark and Performance Evaluation }

\subsubsection{Frame-level Detectors}
As illustrated in Table \ref{tab:benchmark}, we evaluate fifteen representative frame-level detectors  on the GOD-Video dataset. These detectors can be classified into two-stage \cite{ren2015faster,cai2018cascade}, one-stage \cite{liu2016ssd,redmon2018yolov3,lin2017focal,dai2016r}, anchor-free\cite{tian2019fcos,zhang2020bridging,zhou2019objects,dong2020centripetalnet}, and DETR-based detectors \cite{carion2020end, zhu2020deformable, zhang2022dino }.

Two-stage vs. One-stage. Experimental results demonstrate that two-stage detectors generally outperform one-stage detectors. For example, the multi-stage design of Cascade RCNN \cite{cai2018cascade} substantially improves localization accuracy ($AP_{75}$) and performance on small objects ($AP_{s}$). On the one hand, each frame in GOD-Video dataset typically contains one gaseous object, whereas the COCO dataset has an average of 7.7 object instances per image \cite{lin2014microsoft}. GOD-Video exhibits a more pronounced class imbalance issue, resulting in a scarcity of positive samples during training for one-stage detectors. On the other hand, the ROI pooling or ROI Align \cite{he2017mask} operations in the RPN stage of two-stage detectors enhance the perception of internal object information, which contributes to the weak feature extraction of gaseous objects. 

Anchor-based  vs. Anchor-free. The impact of anchor settings is worth investigating in the GOD task. It can be observed that anchor-free detectors achieve comparable $AP$ to Faster RCNN. The priors provided by anchors help alleviate the difficulty of learning the spatial location of gaseous objects, while the anchor-free design is more suitable to locate objects of different geometries, especially those with rare shapes \cite{duan2020corner}. Among the anchor-free detectors, anchor-point detectors \cite{tian2019fcos,zhang2020bridging} encode and decode object bounding boxes as anchor points with corresponding point-to-boundary distances, while key-point detectors \cite{zhou2019objects,dong2020centripetalnet}  predict the locations of key points using a high-resolution feature map and group these key points to form a box. For the GOD task, anchor-point detectors are a preferable choice compared to key-point detectors due to the intricate and fragile spatial morphology of gaseous objects. This complexity presents challenges when attempting to learn the key-point position directly. 

Dense detectors vs. Sparse detectors. Despite their success on the COCO \cite{lin2014microsoft} benchmark, DETR-based detectors \cite{carion2020end, zhu2020deformable, zhang2022dino } yield suboptimal results in the GOD task. Traditional dense detectors, such as Faster R-CNN, adopt the one-to-many assignment strategy, meaning a ground truth box can have multiple corresponding predicted boxes. In contrast, DETR-based detectors employ the one-to-one assignment strategy, where each ground truth is matched with only one predicted box. We attribute the suboptimal performance of DETR-based detectors to two factors: 1)  Optimization difficulty of one-to-one assignment: The region proposal network in Faster RCNN generates hundreds and thousands of region proposals, with those having an IoU greater than a certain threshold (e.g., 0.5) considered as positive samples. The densely generated boxes, with hand-crafted positive and negative sample assignment, provide more high-quality region proposals to the detection heads. Conversely, sparse detectors employ a fixed small set of learned proposal boxes. These sparsely distributed boxes do not cover the ground truth as comprehensively as dense detectors in the initial stage, making the training of sparse detectors more challenging. 2) Saliency deficiency and indistinct boundaries: Conditional DETR \cite{meng2021conditional} reveals that the cross-attention head in the DETR decoder aims to localize four object extremities through the interaction of object queries and image context. However, due to the inherent characteristics of gaseous objects, such as saliency deficiency and indistinct boundaries, the cross attention decoder in sparse detectors struggles to find object extremities for localization during the training. The above experiments provide new insights for the selection of object detectors. When objects to be detected lack salient visual features or distinct boundaries, sparse detectors may exhibit suboptimal performance.

The performance can be enhanced by addressing the class imbalance problem. For example, RetinaNet \cite{lin2017focal} utilizes focal loss to pay attention on the more challenging samples, resulting in the $AP_{50}$ of 37.13\%. ATSS \cite{zhang2020bridging} automatically selects positive and negative samples based on the statistical characteristics of objects, improving $AP_{50}$ from 34.78\% to 37.44\%. This underscores the significance of defining positive and negative samples during training. We believe that the comparison of  frame-level detectors can provide guidance for future improvements in the GOD task.

\begin{table*}[ht]

\caption{The ablation study of the improvement from Faster RCNN to VSF RCNN. The video-level Faster RCNN baseline employs concatenation operations to leverage temporal information. The $VSF_{data}$ and $VSF_{fea}$ are incorporated into the baseline to construct the VSF RCNN. }
\vspace{-1.5em}
\begin{center}
\renewcommand{\arraystretch}{1.2}
\setlength{\tabcolsep}{3.4mm}{
\begin{tabular}{c|c|cccccccc}
\hline
                             & Methods                                                           & $AP_{50}$  & $AP_{75}$  & $AP_{clear}$ & $AP_{vague}$ & $AP_{s}$ &$AP_{m}$& $AP_{l}$& $AP$   \\ \hline
Frame-level                  & Faster RCNN \cite{ren2015faster}                                                       & 36.15 & 6.82  & 17.41     & 9.60      & 3.88  & 8.28  & 22.27 & 13.08 \\ \hline
\multirow{3}{*}{Video-level} & Baseline ($Concat$)                                                            & 37.34 & 8.48  & 20.31     & 9.50      & 5.00  & 9.94  & 23.68 & 14.43 \\
                             &Baseline (+$VSF_{data})$                                                      & 43.88 & 10.90 & 25.17     & 10.71     & 5.74  & 12.12 & 28.22 & 17.35 \\
                             & VSF RCNN (+$VSF_{data\&fea})$ 
  & 51.08 & 12.97 & 28.99     & 13.02     & 6.67  & 15.17 & 31.56 & 20.43 \\ \hline
\end{tabular}}
\end{center}
\label{tab:vsf}
\vspace{-0.0em}
\end{table*}

\subsubsection{Video-level Detectors}
For the video-level detectors, we compare eight representative methods from the tasks of video object detection and spatio-temporal action detection.

Video Object Detection Methods. We choose the DFF \cite{zhu2017deep}, FGFA \cite{zhu2017flow}, SELSA \cite{wu2019sequence} and TRA \cite{wu2019sequence}, which are all built on the Faster RCNN \cite{ren2015faster} detectors for fair evaluation. DFF and FGFA  utilize the guidance of flow estimation to align and warp adjacent features, benefiting from the widely adopted optical flow technique in video analysis known for its effectiveness in exploiting temporal information. However, in the context of the GOD task, the successful extraction of optical flow information is contingent on a high concentration of gaseous objects. Misguided optical flow information can considerably impair the accuracy of DFF and FGFA, which is lower than the Faster RCNN baseline. Therefore, the pre-extraction of optical flow for the GOD task is deemed impractical, particularly considering its heavy reliance on texture and color features in the spatial domain. SELSA employs a comprehensive sequence-level approach for feature aggregation, whereas TRA introduces the temporal ROI align operator to capture temporal information from the entire video regarding  ongoing proposals. Nevertheless, they merely achieve minimal performance improvements compared to Faster RCNN. We contend that the basis for propagating information across frames in video object detection lies in the presence of keyframes that exhibit discriminative and robust features. Unfortunately, this assumption does not hold true for the GOD task.

Spatio-temporal Action Detection Methods. MOC \cite{li2020actions} first extracts features for each frame with DLA34 \cite{yu2018deep}, which is pretrained on COCO dataset to enhance its spatial representation capability.  These features are then concatenated to compose the video feature map. Nevertheless, the concatenation-based design lacks in-depth mining of temporal information. Moreover, the inclusion of additional optical flow results in a decrease in  $AP_{50}$ from 36.81\% to 34.50\%. This observation suggests that the dense-flow, extracted using an external off-the-shelf method \cite{brox2004high}, is ineffective in capturing motion estimation of gaseous objects, which is consistent with the experimental results of DFF \cite{zhu2017deep} and FGFA \cite{zhu2017flow}. Context-aware RCNN \cite{wu2020context} demonstrates only minimal improvements, as it still depends on single-frame detectors for localization. Meanwhile, the DETR-based STMixer \cite{wu2023stmixer} raises $AP_{50}$ from 26.41\% to 32.60\%, showing a comparative enhancement over DETR. However, similar to frame-level sparse detectors, STMixer exhibits inferior performance in the GOD task compared to video-level dense detectors.

In conclusion, previous video-level detectors have predominantly relied on well-established frameworks to generate region proposals. These detectors prioritize modeling the temporal relationships after localization in order to attain a more comprehensive understanding of the video content. In the context of gaseous object detection, it is crucial to pay more attention to the collaborative spatio-temporal representation during the region proposal stage. This is particularly necessary for overcoming the limitation associated with saliency deficiency in the spatial dimension.

\begin{figure}
	\begin{center}
		\includegraphics[width=0.9\linewidth]{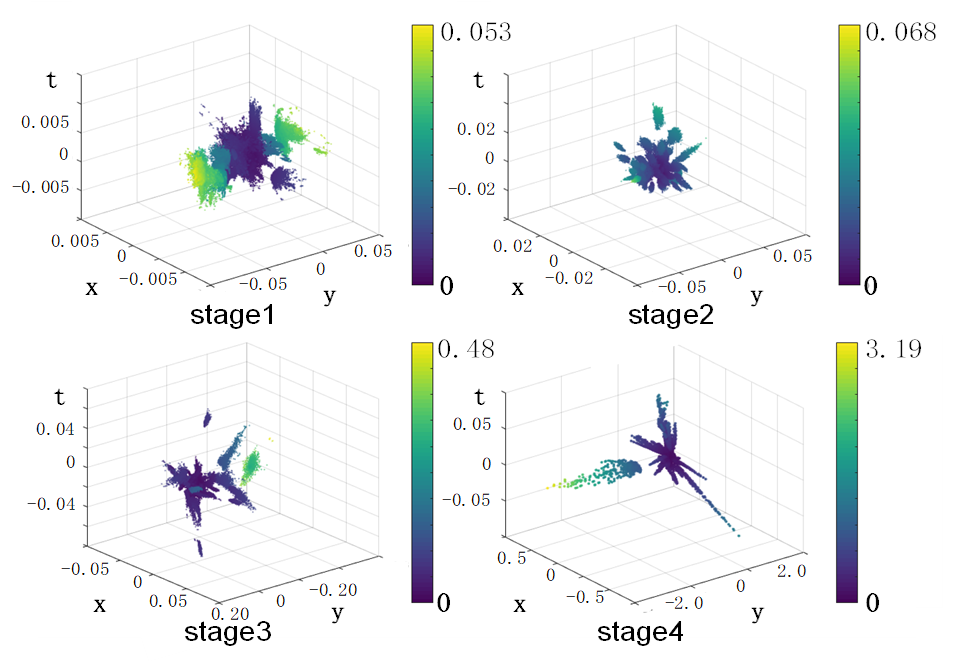}
	\end{center}
	\vspace{-1.0em}
	\caption{The visualization of the learnable offsets at different stages in the ResNet50 \cite{he2016deep}.  }
	\label{fig:vsf_vis}
		\vspace{-0.0em}
\end{figure}

\subsection{Effectiveness of VSF}
We introduce the voxel shift field to facilitate the spatio-temporal feature representation and overcome the limited image cue in a single frame. To evaluate the effectiveness of VSF, we conduct an assessment using the classical Faster RCNN as the frame-level and video-level baseline. For the video-level baseline, we employ concatenation to combine features of multiple frames, as commonly done in previous methods. However, as shown in Table \ref{tab:vsf}, simple concatenation has limited benefits for detecting gaseous objects. Compared to the frame-level Faster RCNN, the video-level concatenation only improves the baseline $AP_{50}$ from 36.15\% to 37.34\%. The poor performance of concatenation  can be attributed to the inconspicuousness of gaseous objects  in single frames and  the similarity between consecutive frames. Consequently, traditional methods of extracting features independently per frame and then concatenating them are insufficient for capturing temporal correlations. Therefore, we believe that the concatenation operation is more suitable for fusing  multiple frames with distinct features.


\begin{table*}[t]
\caption{The representative action recognition methods are selected from the state-of-the-arts on Sth-Sth dataset \cite{goyal2017something},
the spatio-temporal behaviors of VSF and action recognition methods in the GOD task are compared in a unified framework.}
\vspace{-1.5em}
\begin{center}
\renewcommand{\arraystretch}{1.2}
\setlength{\tabcolsep}{4.0mm}{
\begin{tabular}{c|c|cccccccc}
\hline
                                & Methods& $AP_{50}$  & $AP_{75}$  & $AP_{clear}$ & $AP_{vague}$ & $AP_{s}$ &$AP_{m}$& $AP_{l}$& $AP$   \\ \hline
Concat                          & Concat  & 43.88 & 10.90  & 25.17     & 10.71      & 5.74  & 12.12  & 28.22 & 17.35 \\ \cline{1-1}
\multirow{2}{*}{3D-Convolution} & I3D \cite{carreira2017quo} & 47.15 & 11.46 & 26.05     & 12.14     & \textbf{7.17}  & 13.54 & 28.89 & 18.58 \\
                                & S3D \cite{xie2018rethinking} & 47.44 & 10.41 & 25.96     & 11.89     & 6.67  & 12.47 & 29.49 & 18.30 \\ \cline{1-1}
Flow-based                      & MSNet \cite{kwon2020motionsqueeze} & 47.66 & 11.98 & 26.79     & 12.07     & 6.20  & 13.92 & 29.61 & 18.91 \\ \cline{1-1}
Temporal Difference             & TDN \cite{wang2021tdn}  & 46.65 & 10.44 & 25.40     & 11.86     & 5.99  & 12.07 & 29.52 & 18.01 \\ \cline{1-1}
\multirow{3}{*}{Temporal Shift} & TAM \cite{fan2019more} & 48.22 & 11.69 & 26.89     & 12.14     & 6.32  & 13.92 & 29.65 & 18.96 \\
                                & TIN \cite{shao2020temporal} & 47.75 & 11.13 & 26.36     & 12.02     & 5.43  & 12.84 & 30.08 & 18.55 \\
                                & TSM \cite{lin2019tsm} & 48.16 & 11.36 & 26.43     & 12.42     & 6.15  & 12.92 & 30.26 & 18.76 \\ \hline
Ours                            & VSF     & \textbf{51.08} & \textbf{12.97 }& \textbf{28.99 }    & \textbf{13.02}     & 6.67  &\textbf{15.17}& \textbf{31.56} & \textbf{20.43} \\ \hline
\end{tabular}}
\end{center}
\label{tab:action}
\vspace{-0.0em}
\end{table*}
The voxel shift field, specifically designed for the GOD task, overcomes these limitations from both the data and feature aspects. It treats multiple-frame features as a whole for spatio-temporal representation instead of handling them independently. Experimental results demonstrate that $VSF_{data}$, a simple yet powerful operation, increases $AP_{50}$ from 37.34\% to 43.88\%. Additionally, $VSF_{fea}$ is designed to model geometric irregularities and captures the ever-changing shapes in the potential 3D space. The combination of $VSF_{data}$ and $VSF_{fea}$ further improves $AP_{50}$ to 51.08\% and $AP$ to 20.43\%. As illustrated in Fig.~\ref{fig:vsf_vis}, we conduct visualization of the learnable offset at different stages during the feature extraction process. Our observations indicate that in the initial stage, the shift values are relatively small. As the network stage deepens, the VSF progressively captures larger receptive fields, thereby allowing for the extraction of more comprehensive information in the potential space. Additionally, the shift direction becomes more discernible, highlighting the effectiveness of the VSF in capturing spatial-temporal relationships. These findings indicate that VSF can impart the conventional 2D detector, Faster RCNN, with 3D perception capability, thereby establishing a simple yet robust baseline for the GOD task.

\subsection{Spatio-temporal Backbone Analysis}
We systematically compare several representative action recognition models to understand their spatio-temporal behaviors in the GOD task. Specifically, all these models are built on the ResNet50 \cite{he2016deep} backbone, and the input data undergo processing through $VSF_{data}$, which ensures the consistency and comparability across different methods in a unified Faster RCNN framework. The action recognition models are selected from the state-of-the-arts on Sth-Sth dataset \cite{goyal2017something}, which prioritizes motion over scene-focused aspects. As illustrated in Table \ref{tab:action}, I3D \cite{carreira2017quo} serves as a widely adopted and fundamental 3D convolution model, exhibiting reasonably good performance on the $AP$ metric. However, it falls short in explicit mining of temporal relationships and imposes a heavy computational burden. The trainable module proposed in MSNet \cite{kwon2020motionsqueeze} showcases strong performance in the GOD task by establishing correspondences across multiple frames and transforming them into motion features. Implicitly extracting motion information may alleviate the challenge of directly extracting dense optical flow. Models based on the temporal shift \cite{fan2019more,shao2020temporal,lin2019tsm}, where specific channels are shifted along the temporal dimension, demonstrate the ability to achieve superior performance. In our perspective, temporal shift models maintain the integrity of spatial features during the interaction of temporal information. Nevertheless, temporal shift models are primarily limited to shifting along the temporal axis alone, whereas our voxel shift field significantly increases the degree of freedom by enabling shifts  in both spatial and temporal dimensions. Another crucial distinction lies in the implementation of shift operation at the voxel level, as opposed to the conventional channel level. By incorporating a flexible and learnable shifting strategy, the VSF establishes a solid baseline, yielding an $AP_{50}$ of 51.08\% and an $AP$ of 20.43\%.

\subsection{Ablation Studies}
\subsubsection{Applied Stages}
Experimental results presented in Table \ref{tab:stage} show that the application of VSF in the initial stage primarily yields the improvement for $AP_{50}$.  With the progressive application of the voxel shift field in deeper stages, it exhibits an enhanced ability to model robust 3D correlations and facilitate interactions among longer sequences. Consequently, this advancement contributes to a continual improvement in the performance of VSF RCNN, showcasing its effectiveness in capturing complex spatio-temporal patterns.

\begin{table}[t]
\caption{Ablation study of the voxel shift field with different applied stages.}
\vspace{-1.5em}
\begin{center}
\renewcommand{\arraystretch}{1.2}
\setlength{\tabcolsep}{1.9mm}{
\begin{tabular}{cclc|ccccc}
\hline
\multicolumn{4}{c|}{Stage}        & \multirow{2}{*}{$AP_{50}$} & \multirow{2}{*}{$AP_{75}$} & \multirow{2}{*}{$AP_{clear}$} & \multirow{2}{*}{$AP_{vague}$} & \multirow{2}{*}{$AP$} \\ \cline{1-4}
1 & 2 & 3                     & 4 &                       &                       &                            &                            &                     \\ \hline
&                          &   &          & 43.88   &   10.90 & 25.17    &  10.71         &   17.35    \\
$\checkmark$  &   &                       &   &  45.22   &   11.11 & 25.99    &  11.21         &   17.95  \\
$\checkmark$  & $\checkmark$  &                       &   &             47.71   &   11.95 & 27.25    &  11.99         &   19.04              \\
$\checkmark$  & $\checkmark$ & \multicolumn{1}{c}{$\checkmark$ } &   &            50.81   &   12.56 & 28.99    &  12.85         &   20.31                   \\ \hline
$\checkmark$  & $\checkmark$ & \multicolumn{1}{c}{$\checkmark$ } &$\checkmark$  & 51.08 & 12.97 & 28.99     & 13.02     & 20.43         \\ \hline
\end{tabular}}
\end{center}
\vspace{-1.5em}
\label{tab:stage}
\end{table}

\begin{table}[t]
\caption{Ablation study of the voxel shift field with different configurations of the shift direction.}
\vspace{-1.5em}
\begin{center}
\renewcommand{\arraystretch}{1.2}
\begin{tabular}{ccc|ccccc}
\hline
\multicolumn{3}{c|}{Shift Direction} & \multirow{2}{*}{$AP_{50}$} & \multirow{2}{*}{$AP_{75}$} & \multirow{2}{*}{$AP_{clear}$} & \multirow{2}{*}{$AP_{vague}$} & \multirow{2}{*}{$AP$} \\ \cline{1-3}
W          & H          & T          &                       &                       &                            &                            &                     \\ \hline
       &            &          & 43.88   &   10.90 & 25.17    &  10.71         &   17.35                  \\
$\checkmark$          &            &            &        47.28      &     11.54 &       27.08       &    11.65               &    18.70               \\
           & $\checkmark$          &            &                    46.90      &     11.40 &       26.85       &    11.61               &    18.58               \\
$\checkmark$        &$\checkmark$          &     &    48.18  &  11.66                     & 27.53  &  11.90  &     19.07   \\
           &            & $\checkmark$          &       50.33    &     12.11       &     28.15       &                 12.74      &       19.89      \\ \hline
$\checkmark$     & $\checkmark$        & $\checkmark$          &  51.08 & 12.97 & 28.99     & 13.02     & 20.43              \\ \hline
\end{tabular}
\end{center}
\vspace{-1.5em}
\label{tab:direction}
\end{table}

\begin{table}[t!]
\caption{Ablation study of the voxel shift field with different initial biases.}
 \vspace{-1.5em}
\begin{center}
\renewcommand{\arraystretch}{1.2}
\begin{tabular}{clll|ccccc}
\hline
\multicolumn{4}{c|}{Initial Bias} & $AP_{50}$ & $AP_{75}$ &$AP_{clear}$ &$AP_{vague}$ & $AP$  \\ \hline
\multicolumn{4}{c|}{$\pm$1}       & 50.61   &    12.79      &  28.90         &   12.79   & 20.24        \\
\multicolumn{4}{c|}{$\pm$2}    & 50.51  &  12.47   &   28.97  & 12.69  &    20.13   \\
\multicolumn{4}{c|}{$\pm$3}    & 50.54  &  12.07  &   28.59  & 12.56  &    19.91    \\
\multicolumn{4}{c|}{$\pm$1 $\pm$2}            & 51.08  & 12.97  &   28.99   &  13.02  &  20.43\\
\multicolumn{4}{c|}{$\pm$1 $\pm$3}        & 50.82 & 12.56 & 28.69     & 12.72     & 20.12    \\
\multicolumn{4}{c|}{$\pm$1 $\pm$2 $\pm$3}           & 51.27 & 12.75 &  28.74 &  12.99 &  20.31\\ \hline
\end{tabular}
\end{center}
 \vspace{-0.0em}
 \label{tab:number}
 \label{bias}
\end{table}

\subsubsection{Shift Directions}
To assess the influence of shift directions in the voxel shift field, we conduct ablation experiments concentrating on different shifting directions as detailed in Table \ref{tab:direction}. Initially, we restrict the freedom of shift solely to the height (H) and width (W) dimensions. This restriction yields a minor improvement compared to the baseline. Subsequently, when both the height and width dimensions are considered simultaneously, limiting the shift to spatial dimensions, an increase of 4.30\% in $AP_{50}$ is observed. Furthermore, the inclusion of shift in the temporal dimension (T) shows significant benefits, resulting in a substantial 6.45\% improvement in $AP_{50}$. This finding suggests that temporal dimension plays a more critical role in the GOD task, underscoring the significance of allocating more attention.

\begin{table}[t]
\caption{Evaluation results with different input frame numbers.}
 \vspace{-1.5em}
\begin{center}
\renewcommand{\arraystretch}{1.2}
\begin{tabular}{clll|ccccc}
\hline
\multicolumn{4}{c|}{Frame Number} & $AP_{50}$ & $AP_{75}$ &$AP_{clear}$ &$AP_{vague}$ & $AP$  \\ \hline
\multicolumn{4}{c|}{1}       & 36.15   &    6.82  & 17.41     &  9.60         &   13.08            \\
\multicolumn{4}{c|}{2}    & 41.68  &  9.34 &    22.88   &   10.29  & 15.99    \\
\multicolumn{4}{c|}{4}            & 47.87  & 11.39  &   27.09   &  11.78  &  18.74\\
\multicolumn{4}{c|}{8}        & 51.08 & 12.97 & 28.99     & 13.02     & 20.43    \\
\multicolumn{4}{c|}{16}           & 51.23 & 12.60 &  29.13 &  12.84 &  20.40\\ \hline
\end{tabular}
\end{center}
 \vspace{-1.5em}
 \label{tab:number}
\end{table}

\begin{table}[t]
\caption{The effectiveness validation of VSF based on other detectors.}
\vspace{-1.5em}
 \begin{center}
  \renewcommand{\arraystretch}{1.2}
  \setlength{\tabcolsep}{2.0mm}{
\begin{tabular}{clll|ccccc}
\hline
\multicolumn{4}{c|}{Detector}      & \multicolumn{1}{c}{$AP_{50}$} & \multicolumn{1}{c}{$AP_{75}$} & \multicolumn{1}{c}{$AP_{clear}$} & \multicolumn{1}{c}{$AP_{vague}$} & \multicolumn{1}{c}{$AP$} \\ \hline
\multicolumn{4}{c|}{RetinaNet \cite{lin2017focal}}           & 37.13                     & 7.23                      & 18.70                             & 9.56                             & 13.51                  \\
\multicolumn{4}{c|}{$Concat$}        & 38.38                     & 7.89                      & 19.82                             & 10.21                            & 14.48                  \\
\multicolumn{4}{c|}{+$VSF_{data}$}      &         42.55          &        7.98      &          22.20         &       10.41              &    15.61       \\
\multicolumn{4}{c|}{+$VSF_{data\&fea}$} & 50.35                     & 10.57                     & 26.35                             & 13.04                            & 19.05                  \\ \hline
\multicolumn{4}{c|}{FCOS \cite{tian2019fcos} }         & 34.78                     & 6.80                      & 17.25                             & 8.86                             & 12.64                  \\
\multicolumn{4}{c|}{$Concat$}        & 36.49                     & 7.38                      & 18.93                             & 8.87                             & 13.38                  \\
\multicolumn{4}{c|}{+$VSF_{data}$}      & 40.45                     & 7.10                      & 20.88                             & 9.02                             & 14.40                  \\
\multicolumn{4}{c|}{+$VSF_{data\&fea}$} & 48.75                    & 9.50                      & 25.57                             & 11.25                           & 17.90                  \\ \hline
\multicolumn{4}{c|}{DETR \cite{carion2020end} }         & 26.41                     & 3.75                      & 12.90                             & 5.62                             & 8.66                  \\
\multicolumn{4}{c|}{$Concat$}        & 32.97                     & 5.40                      & 15.50                             & 8.16                             & 11.45                  \\
\multicolumn{4}{c|}{+$VSF_{data}$}      & 37.25                     & 5.24                      & 18.26                             & 7.50                             & 12.49                  \\
\multicolumn{4}{c|}{+$VSF_{data\&fea}$} & 40.96                    & 6.48                      & 20.91                             & 8.63                           & 14.18                  \\ \hline

\end{tabular}}
 \end{center}
 \vspace{-0.5em}
 \label{tab:fcos}
\end{table}

\subsubsection{Initial Biases}
The design of bias terms aims to facilitate interaction between different temporal information through initial biases along the t-axis. We conduct additional ablation experiments on bias settings, as presented in Table~\ref{bias}. In the ablation experiment with a single bias term, the bias set of $\pm$1 achieves the best performance. As the bias increases from $\pm$1 to $\pm$3, the $AP$ gradually decreases  from 20.24\% to 19.91\%. The combination of bias $\pm$1 and $\pm$2 is able to further improve the $AP$ to 20.43\%. Nevertheless, introducing an additional $\pm$3 bias  will increase the $AP_{50}$ but decrease the $AP$. In our view, the increase in the bias set will expand the temporal interaction range, potentially leading to ambiguities in identifying the boundaries of gaseous objects. Such ambiguities result in reduced precision under higher IoU evaluation metrics (e.g., $AP_{75}$), thereby diminishing the overall $AP$. Table~\ref{bias} shows that the combination of bias ±1 and ±2 yields the best results. Therefore, we have adopted this initial bias configuration.

\begin{figure*}
	\begin{center}
		\includegraphics[width=1.0\linewidth]{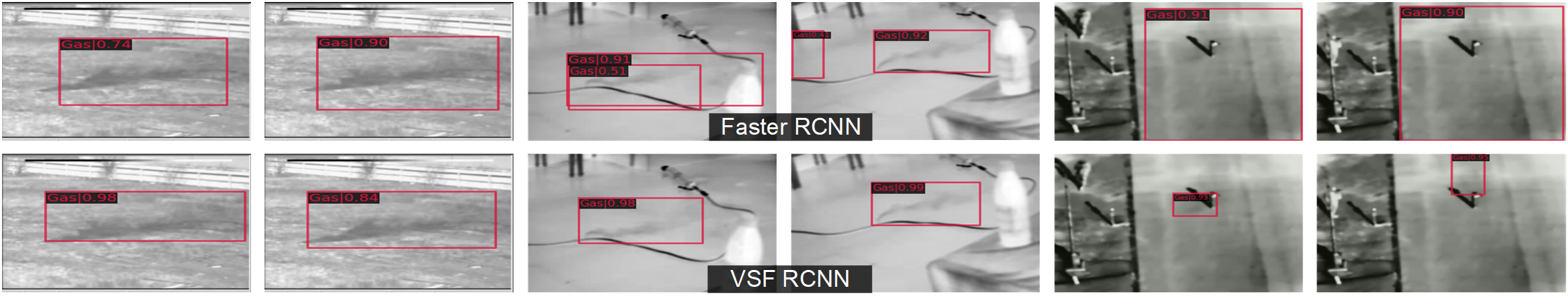}
	\end{center}
	\vspace{-2.0em}
	\caption{ In comparisons of the Faster RCNN and VSF RCNN on three representative samples, VSF RCNN exhibits a superior detection performance. }
	\label{fig:vsfrcnn_vis}
		\vspace{-1.0em}
\label{fig:faster_vsf}
\end{figure*}

\begin{figure*}
	\begin{center}
		\includegraphics[width=1.0\linewidth]{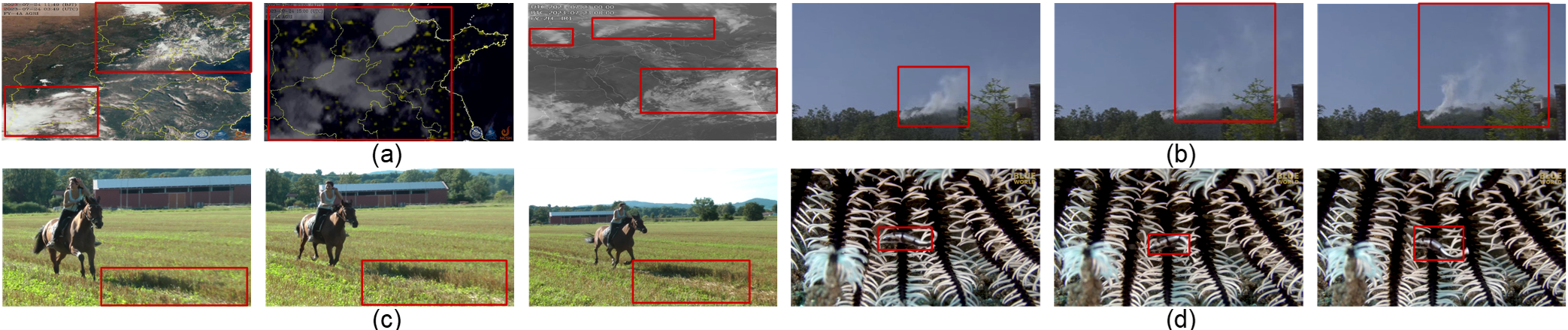}
	\end{center}
	\vspace{-1.0em}
	\caption{ We investigate the performance of VSF RCNN in the related tasks without fine-tuning or retraining. (a) Meteorological satellite cloud image processing\protect\footnotemark[1]. (b) Video smoke detection \cite{ko2011modeling}. (c) Video shadow detection \cite{lu2022video}. (d) Video camouflaged object detection \cite{lamdouar2020betrayed}. As VSF RCNN is trained with the infrared modality, the samples in the aforementioned tasks are converted to grayscale to ensure a similar image style. }
	\label{fig:app_vis}
		\vspace{-0.0em}
\label{fig:vsf_app}
\end{figure*}

\subsubsection{Different Input Frames}
 The range of input frames is considered crucial in capturing temporal information. Therefore, we assess the performance of our VSF RCNN with different sets of input frames, as presented in Table \ref{tab:number}. Intuitively, increasing the number of input frames from 1 to 8 would lead to a higher average precision. Nevertheless, the  performance of VSF RCNN with 16 input frames is comparable to 8 input frames. This finding suggests that our baseline method primarily emphasizes short-term temporal modeling. It highlights the necessity for improved long-term temporal modeling capabilities with novel insights and underscores the potential for continued exploration and refinement of our approach. 

\subsubsection{Applicability on Other Detectors}
 To demonstrate the applicability of the voxel shift field on other 2D detectors, we incorporate it into  various detectors for effectiveness validation. The implementation is aligned with VSF RCNN to ensure a fair comparison. Specifically, VSF is inserted into each ResNet50 at the backbone level,  and a detection head responsible for classification and regression tasks is independently constructed for each frame. Experimental results presented in  Table \ref{tab:fcos} demonstrate  remarkable performance improvements for FCOS \cite{tian2019fcos},  RetinaNet \cite{lin2017focal}, and DETR \cite{carion2020end} with  the incorporation of VSF. These findings confirm the generalizability of our voxel shift field to mainstream 2D detectors, including two-stage, single-stage, anchor-free, and DETR-based detectors. The VSF can be seamlessly integrated into the majority of established detectors, imparting conventional 2D detectors with 3D modeling capabilities.


\subsection{Visualization Results}
Fig.~\ref{fig:faster_vsf} provides a visualization of the detection results between Faster RCNN and VSF RCNN, focusing on three representative samples. The first sample involves a clear gas leak scene, where VSF RCNN demonstrates superior performance in precise boundary regression compared to Faster RCNN. In the second sample, which showcases a rapidly moving gaseous object along pipes, Faster RCNN generates multiple mismatched boxes, while VSF RCNN accurately tracks the location of the gaseous object in time. In the third sample, where the gas is barely visible in a single frame image, Faster RCNN erroneously identifies the entire ground as a predicted box. Contrarily, VSF RCNN effectively detects the faint leak in the distant vicinity by leveraging efficient spatio-temporal representation capabilities. These visual comparisons illustrate the significant advantages offered by VSF RCNN in handling challenging scenarios with improved accuracy and robustness.


In addition, we investigate the performance of VSF RCNN in various related tasks. Without the need for fine-tuning or retraining, we directly employ the trained model of VSF RCNN  for tasks involving meteorological satellite cloud image processing, video smoke detection \cite{ko2011modeling}, video shadow detection  \cite{lu2022video} , and video camouflaged object detection \cite{lamdouar2020betrayed}. It is worth noting that these tasks involve RGB images, whereas  VSF RCNN is initially trained on the infrared modality. To address this discrepancy, we adopt a straightforward approach that converts a sequence of eight consecutive RGB images into grayscale images, which served as the input for VSF RCNN. The similarity between the cloud in Fig.~\ref{fig:vsf_app} (a) and the smoke in Fig.~\ref{fig:vsf_app} (b), as well as the gaseous object in GOD-Video, allows VSF RCNN to yield satisfactory localization outcomes across these diverse tasks. VSF RCNN also showcases its capabilities in detecting shadow objects in Fig.~\ref{fig:vsf_app} (c), which exhibit irregular shapes and lower brightness levels compared to their surroundings. Moreover, as depicted in Fig.~\ref{fig:vsf_app} (d), the small fish is challenging to detect in a single frame due to its concealed appearance. However, by leveraging its robust temporal modeling capability, VSF RCNN tracks the movement of the object well throughout the video sequence. The aforementioned results demonstrate the versatility and adaptability of VSF RCNN across diverse application domains, exhibiting its capability to irregular geometry shapes in videos. In addition, the investigation conducted by GOD can serve as a catalyst for similar research endeavors.
\footnotetext[1]{The video samples are from the National Satellite Meteorological Center website: \url{http://www.nsmc.org.cn/nsmc/en/home/index.html} }

\subsection{Error Analysis}

\begin{figure*}[t]
	\begin{center}
		\includegraphics[width=1.0\linewidth]{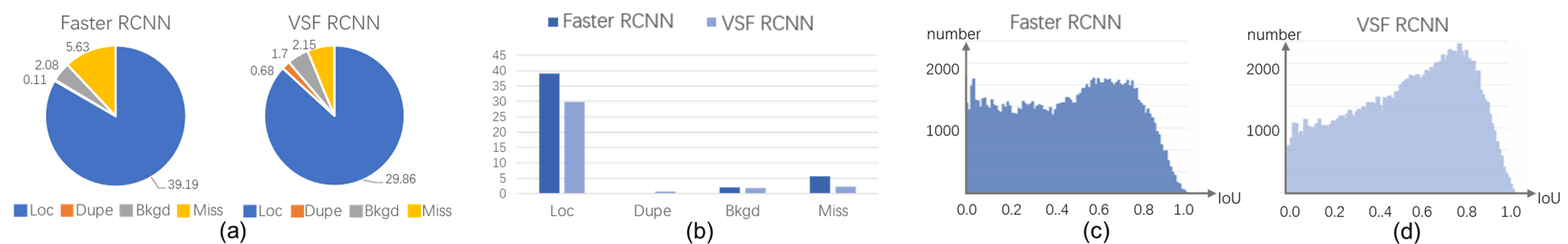}
	\end{center}
	\vspace{-1.0em}
	\caption{ Error analysis with the TIDE toolbox \cite{bolya2020tide}. (a) The pie chart shows the relative contribution of each error. (b) The bar plots show the absolute contribution of each error. (c-d)  The statistical distribution of IoU density for Faster RCNN and VSF RCNN in the GOD-Video test set. }
	\label{fig:app_vis}
		\vspace{-0.0em}
\label{fig:tide}
\end{figure*}

We adopt TIDE, a general toolbox for identifying object detection errors \cite{bolya2020tide}, to analyze the failure cases of Faster RCNN and VSF RCNN.
Since the GOD-Video dataset contains only one category, we focus on four specific error types in the TIDE: localization error (Loc), duplicate detection error (Dupe), background error (Bkgd), and missed GT error (Miss). Fig.~\ref{fig:tide} (a) presents the pie chart showing the relative contributions of each error type, indicating that localization error is the primary error source. Fig.~\ref{fig:tide} (b) compares the absolute contribution of different error types between Faster RCNN and VSF RCNN, which demonstrates that VSF RCNN significantly reduces both localization error and missed GT error. To conduct a further analysis of the localization error, we calculate the IoU between the predicted box with the highest score and the ground truth.  Fig.~\ref{fig:tide} (c-d) shows the statistical distribution of IoU density within the GOD-Video test set. It is evident that VSF RCNN exhibits a higher distribution in the high IoU region compared to Faster RCNN. The voxel shift field mitigates  the impact of boundary ambiguity by jointly extracting spatio-temporal features, thereby improving the IoU between predicted boxes and the ground truth.

\section{POTENTIAL RESEARCH DIRECTIONS}
Considering gaseous object detection is an emerging field, we propose several potential directions for future research:

1) Extensibility of VSF.  We believe that the voxel shift field holds potential for versatile application across a spectrum of tasks that require multi-dimensional feature extraction. These tasks involve the spatio-temporal dimension, such as video smoke detection \cite{ko2011modeling} and camouflage video object detection \cite{lamdouar2020betrayed}; the three-dimensional space, as seen in 3D point cloud object detection \cite{deng2021voxel} and pulmonary nodule detection \cite{mei2021sanet}; and the spatial-spectral dimension, exemplified by hyperspectral image processing \cite{khan2018modern}. Furthermore, VSF can also be explored in low-level vision tasks, such as video-level denoising \cite{tassano2020fastdvdnet}, super-resolution \cite{yue2022real} and restoration \cite{li2023simple}.

Unlike previous detectors, VSF RCNN trained on GOD-Video demonstrates a unique capability to model the spatio-temporal representation for irregular geometric shapes. We aspire for this capacity to find applicability within scientific research, including but not limited to astronomy, fluid mechanics, atmospheric science \cite{bi2023accurate}, and medical fields, some examples encompass the application of combustion analysis, computer-aided diagnosis of pulmonary nodules \cite{mei2021sanet}  and meteorological satellite cloud image processing.

2) Spatio-temporal Representation. The object detector encompasses several components, including the backbone, head, neck and loss, offering rich opportunities to delve deeper into harnessing temporal information in the GOD task. Furthermore, the transformer architecture is driving a novel paradigm in action recognition \cite{ulhaq2022vision} and holds potential for introduction into the GOD task to enhance the representation of spatio-temporal features.


3) Computational Efficiency. Due to the simultaneous processing of multiple frames in VSF RCNN, the inference speed is significantly slower than Faster RCNN, dropping from around 44.6 FPS to 9.1 FPS on the NVIDIA 3090. Enhancing the computational efficiency of video-level detectors is a promising direction.

4) Gas-centric Vision Task. Based on the GOD-Video dataset, gas-centric visual tasks can be explored such as gaseous object classification, tracking, segmentation and temporal segmentation.

5) Weakly/Semi-Supervised Detection. The annotation of GOD-Video is time-consuming and requires human experts with professional expertise. It is necessary to study the weakly/semi-supervised detection to avoid heavy annotation costs.

6) Advanced Computational Imaging Technology. Recent advancements in computational imaging have enabled more  refined analytical methods. Specifically, hyperspectral gas cloud imaging systems \cite{hagen2013video} facilitate the identification of different gas types. The integration of the GCI system with laser technology, facilitated by multi-sensor fusion, supports the extraction of detailed information such as gas concentration and spatial distances. Additionally, emerging technologies such as single-photon imaging \cite{titchener2022single} open new avenues for further academic exploration.



\section{CONCLUSION}
We present the first comprehensive study on a rarely explored task called gaseous object detection. This task differs significantly from conventional object detection in the following aspects: 1) saliency deficiencies, 2) arbitrary and ever-changing shapes, 3) lack of distinct boundaries. In order to facilitate the study of this challenging task, we construct a large-scale, high-quality, and diversified dataset named GOD-Video. Based on this dataset, we conduct a rigorous evaluation of both frame-level detectors and video-level detectors. Additionally, we develop a voxel shift field to capture geometric irregularities and ever-changing shapes in potential 3D space, thereby enhancing conventional 2D detectors with 3D perceptual capabilities. However, there is still substantial room for improvement in our baseline method, and we identify  several foreseeable directions for future  research. In the future, we plan to combine advanced computational imaging technology with the GOD task to achieve more sophisticated analysis and explore potential applications in scientific research. The proposed GOD-Video dataset and VSF RCNN baseline are expected to attract further research into this valuable albeit challenging task.

\section*{Acknowledgments}
This research was supported by National Science Fund for Distinguished Young Scholars (62025108). We thank the joint  research laboratory of $ZHIPUTECH$ and Nanjing university for the gas imaging camera supports and data collection. 




 
%

\bibliographystyle{IEEEtran}
\bibliography{IEEEabrv}












\newpage
\vspace{-2em}
\begin{IEEEbiography}[{\includegraphics[width=1in,height=1.25in,clip,keepaspectratio]{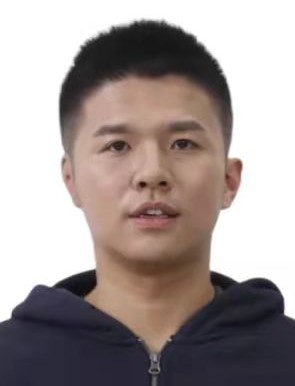}}]{Kailai Zhou} received the BS degree from Dalian University of Technology, Dalian, China, in 2019. He is currently working toward the Ph.D. degree with the School of Electronic Science and Engineering. Nanjing University, Nanjing, China. His research interests include computational spectral imaging and computer vision.
\end{IEEEbiography}
\vspace{-2em}
\begin{IEEEbiography}[{\includegraphics[width=1in,height=1.25in,clip,keepaspectratio]{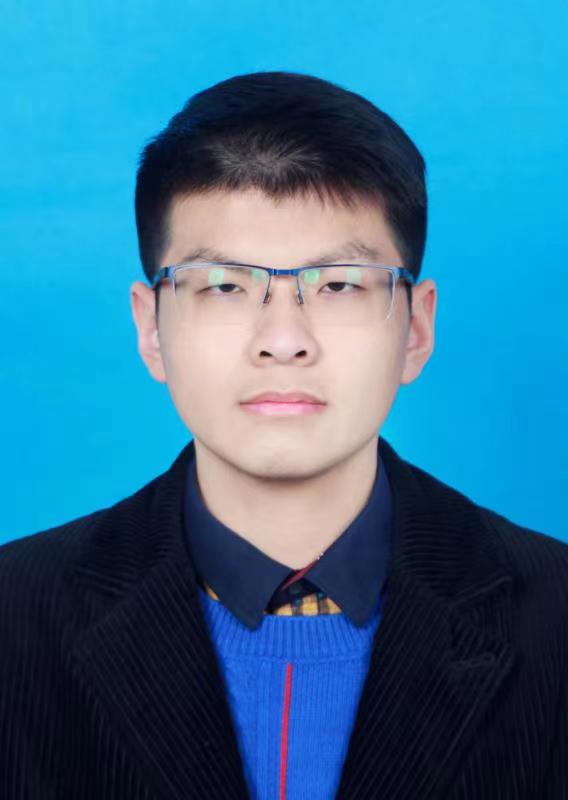}}]{Yibo Wang}  received the BS degree from the College of Electrical and Information Engineering, Hunan University, Changsha, China, in 2021. He is currently working toward the MS degree with the School of Electronic Science and Engineering, Nanjing University, Nanjing, China. His research interests include computer vision and deep learning.
\end{IEEEbiography}

\vspace{-2em}
\begin{IEEEbiography}[{\includegraphics[width=1in,height=1.25in,clip,keepaspectratio]{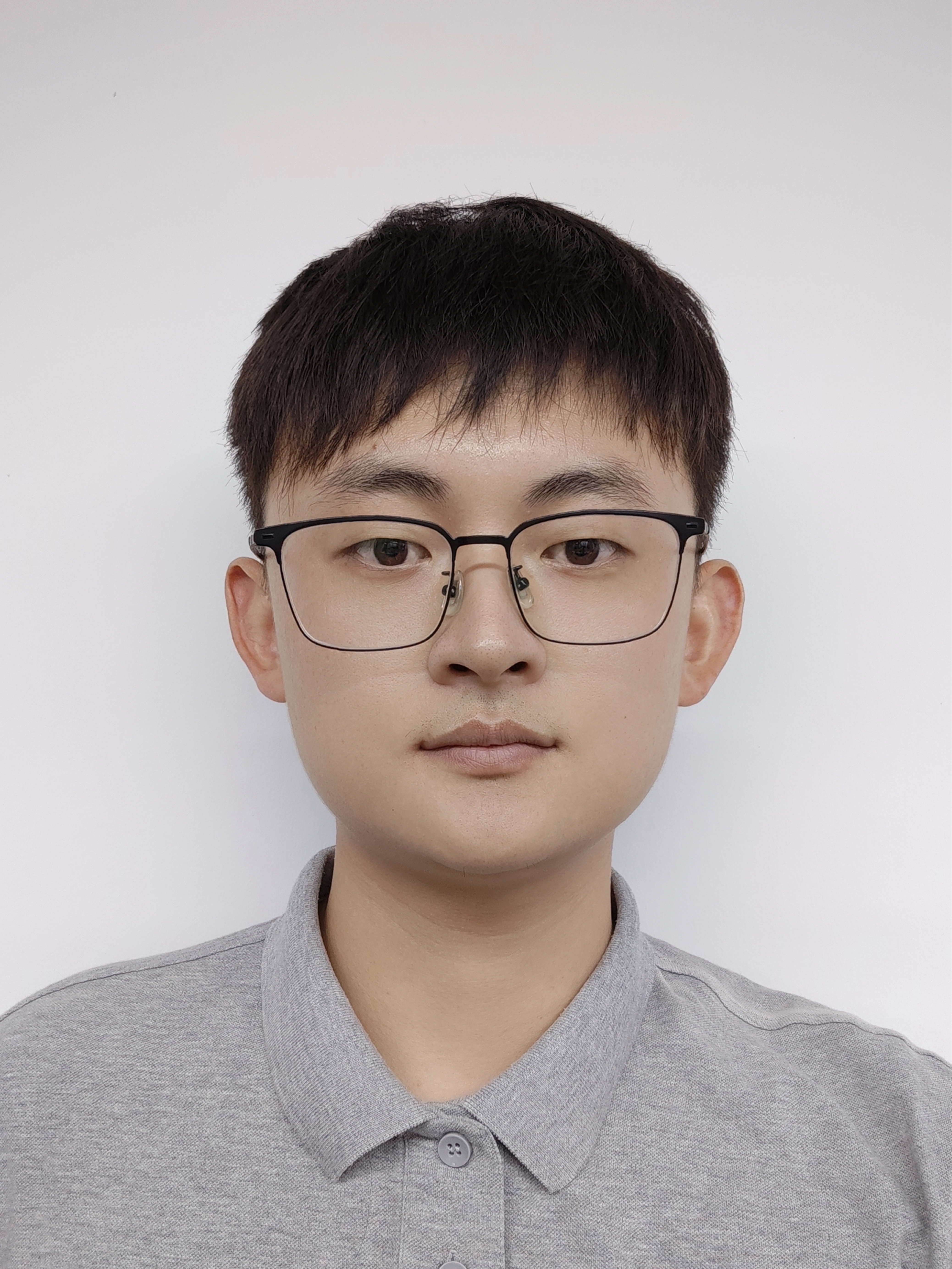}}]{Tao Lv}  received the BS degree from the college of Information Science and Engineering, Northeastern University, Liaoning, China, in 2021. He is currently working toward the Ph.D. degree with the School of Electronic Science and Engineering, Nanjing University, Nanjing, China. His research interests include computational photography and computer vision, especially computational spectral imaging.
\end{IEEEbiography}
\vspace{-2em}
\begin{IEEEbiography}[{\includegraphics[width=1in,height=1.25in,clip,keepaspectratio]{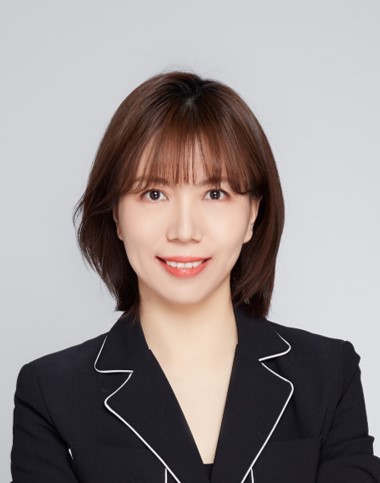}}]{Qiu Shen}  (Member, IEEE) received the BS degree in electrical engineering and information science and PhD degree in signal and information processing from the University of Science and Technology of China, in 2004 and 2009, respectively. From 2009 to 2016, she was with Huawei 2012 Lab and Nanjing University of Aeronautics and Astronautics. She is now on the faculty of Electronic Science and Engineering School, Nanjing University, China. Her current research focuses on computational imaging and computer vision.

\end{IEEEbiography}

\vspace{-2em}
\begin{IEEEbiography}[{\includegraphics[width=1in,height=1.25in,clip,keepaspectratio]{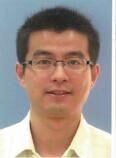}}]{Xun Cao}  (Member, IEEE) received the B.S. degree from Nanjing University, Nanjing, China, in 2006,
and the Ph.D. degree from the Department of Automation, Tsinghua University, Beijing, China, in 2012. He held visiting positions with Philips Research, Aachen, Germany, in 2008, and Microsoft Research Asia,
Beijing, from 2009 to 2010. He was a Visiting Scholar with The University of Texas at Austin, Austin, TX, USA, from 2010 to 2011. He is currently a Professor with the School of Electronic Science and Engineering, Nanjing University. His research interests include
computational photography, image-based modeling and rendering, and VR/AR
systems.
\end{IEEEbiography}

 




\newpage
\appendix

In the supplementary material, we provide more details about the gas imaging model, dataset overview, optical flow and background modeling results, objectness measures, as well as visualization comparisons of detectors.

\section{Gas Imaging Model}
In order to provide a more detailed elucidation of the physical modeling and derivation process of the Gaussian gas dispersion model for voxel shift field, we commence with the gas imaging model, which involves 
the Planck blackbody radiation law, the Lambert-Beer law and the layer radiative transfer model. Based on the above theory, we conduct a thorough analysis of the correlation between the gas dispersion in the real world, the response of the camera, and the voxel shift field in the feature representation stage.
The layer radiative transfer model, commonly referred to as the "Layer Model", stands as one of the extensively employed gas infrared detection radiative transfer models. Under the assumption of excluding environmental radiation and aerosol interference, the expression for the layer radiative transfer model is as follows:
\begin{equation}
\mathrm{M}_{\mathrm{i}}=\tau_{\mathrm{T}} \tau_{\mathrm{A}} \mathrm{M}_{\mathrm{i}-1}+\varepsilon_{\mathrm{T}} \varepsilon_{\mathrm{A}} \mathrm{M}_i^{\mathrm{b}}
\label{eqn:1}
\end{equation}

Here, $M_i$ represents the outgoing radiant flux from the i-th layer, $\tau_\mathrm{T}$ and $\tau_\mathrm{A}$ are the transmittance of the target and the atmosphere for the i-th layer, $\varepsilon_\mathrm{T}$ and $\varepsilon_\mathrm{A}$ are the emissivity of the target and the atmosphere for the i-th layer. $M_i^b$ denotes the blackbody radiation corresponding to the temperature $T_i$ in the i-th layer, and $M_{i-1}$ signifies the outgoing radiant flux from the preceding layer.

\begin{figure*}
	\begin{center}
		\includegraphics[width=1.0\linewidth]{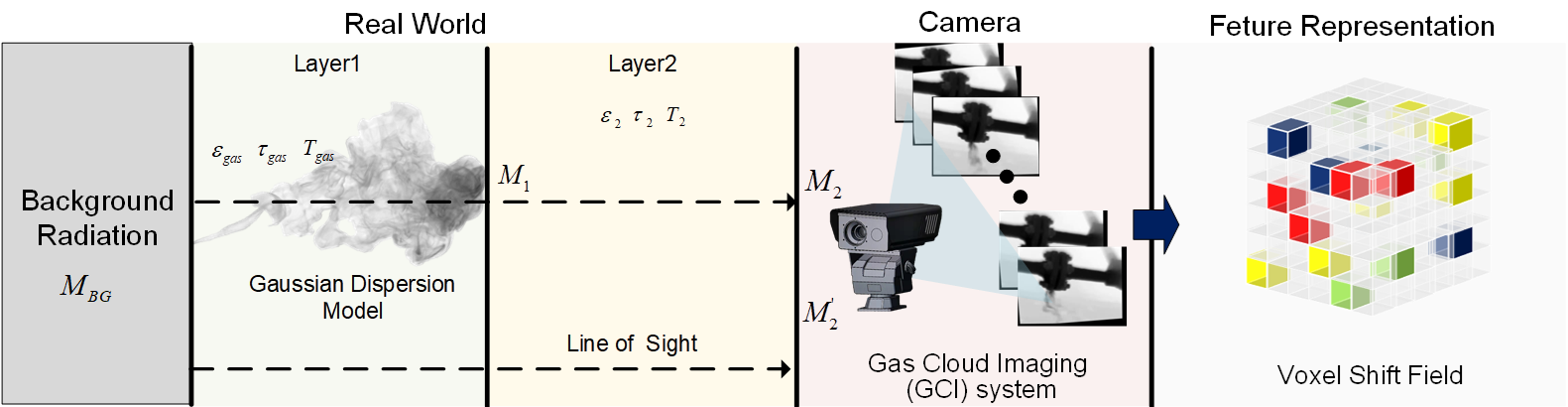}
	\end{center}
	\caption{The schematic diagram of the gas imaging model from the real world to the camera and feature representation stage.
 }
	\label{fig:GIM}
\end{figure*}

To facilitate our research, we assume a uniform distribution of gas or atmospheric components within each gas target layer or atmospheric transmission layer. Additionally, when addressing leakage gas clouds in proximity to the leakage source, buildings, or the ground, we can simplify the layer radiative transfer model into a dual-layer configuration, as illustrated in Fig.~\ref{fig:GIM}.

Within the context of the dual-layer radiative transfer model, there are two distinct paths along the instrument's line of sight (LOS): the gas path (On-plume path) and the non-gas path (Off-plume path). Here, we utilize $M_i$ to represent the total radiation emitted from the i-th layer, while $M_{BG}$ denotes the spectral radiation originating from the background. By applying the layer radiative transfer model from Eq. (\ref{eqn:1}), we can derive the stratified radiative transfer equation for the gas path in layer1 and layer2:
\vspace{-0.3em}
\begin{equation}
\mathrm{M}_1=\tau_{gas}(\lambda) \mathrm{M}_{\mathrm{BG}}\left(\lambda, \mathrm{T}_{\mathrm{b}}\right)+\varepsilon_{gas}(\lambda) \mathrm{M}_{gas}\left(\lambda, \mathrm{T}_{gas}\right) 
\vspace{-1.5em}
\label{eqn:2}
\end{equation}

\begin{equation}
\mathrm{M}_2=\tau_2(\lambda) \mathrm{M}_1\left(\lambda, \mathrm{T}_{\mathrm{b} 1}\right)+\varepsilon_2(\lambda) \mathrm{M}_2\left(\lambda, \mathrm{T}_2\right) 
\label{eqn:3}
\end{equation}

In this context, $\tau_{gas}(\lambda)$ and $\varepsilon_{gas}(\lambda)$ respectively denote the spectral transmittance and emissivity of the gas, while $\tau_2(\lambda)$ and $\varepsilon_2(\lambda)$ represent the spectral transmittance and emissivity of the atmosphere. By substituting the Eq. (\ref{eqn:2}) into Eq. (\ref{eqn:3}), the $M_2$ can be rewritten as:

\begin{equation}
\begin{split}
&\mathrm{M}_2=\tau_2(\lambda) \tau_{gas}(\lambda) \mathrm{M}_{\mathrm{BG}}\left(\lambda, \mathrm{T}_{\mathrm{b}}\right) +   \\
&\tau_2(\lambda) \varepsilon_{gas}(\lambda) \mathrm{M}_{gas}\left(\lambda, \mathrm{T}_{gas}\right)+\varepsilon_2(\lambda) \mathrm{M}_2\left(\lambda, \mathrm{T}_2\right)
\end{split}
\label{eqn:4}
\end{equation}

For the radiative transfer equation concerning the non-gas path, the above expression can be simplified as follows:

\begin{equation}
\mathrm{M}_2^{\prime}=\tau_2(\lambda) \mathrm{M}_{\mathrm{BG}}\left(\lambda, \mathrm{T}_{\mathrm{b}}\right)+\varepsilon_2(\lambda) \mathrm{M}_2\left(\lambda, \mathrm{T}_2\right) 
\end{equation}

Taking into account the equivalence of gas emissivity and absorptivity, and the sum of absorptivity and transmittance is 1, we have $\varepsilon_{gas}(\lambda) = 1 - \tau_{gas}(\lambda)$. Consequently, the expression for the radiance difference between the gas path and the non-gas path can be presented as:

\begin{equation}
\begin{split}
\Delta M &=\mathrm{M}_2^{\prime}-\mathrm{M}_2 \\
&=\tau_2(\lambda)\left[1-\tau_{\text {gas }}(\lambda)\right]\left[\mathrm{M}_{\mathrm{BG}}\left(\lambda, \mathrm{T}_{\mathrm{b}}\right)-\mathrm{M}_{\text {gas }}\left(\lambda, \mathrm{T}_{\text {gas }}\right)\right]
\end{split}
\end{equation}

The background spectral radiance, denoted by $M_{BG}(\lambda,T_b)$, is expressed as follows:
\begin{equation}
M_\mathrm{BG}\left(\lambda, \mathrm{T}_{\mathrm{b}}\right) = \varepsilon_b(\lambda) M\left(\lambda, \mathrm{T}_{\mathrm{b}}\right)
\end{equation}

Where $M(\lambda,T)$ represents the blackbody radiation at temperature T, which follows the specific form of the Planck blackbody radiation formula:

\begin{equation}
\mathrm{M}(\lambda, \mathrm{T}) =\frac{c_1}{\lambda^5} \frac{1}{\exp \left(c_2 / \lambda \mathrm{T}\right)-1} 
\end{equation}
In the equation, $c_1 = 3.74\times 10^{-16} (W \cdot m^2)$ and $c_2 = {1.44\times10}^{-2} (m \cdot K)$ represent the first and second radiation constants, respectively. Similarly, $M_{gas}(\lambda,T_{gas})$ can be expressed using the blackbody radiation at temperature $T_{gas}$.

The spectral transmittance of the gas, $\tau_{gas}(\lambda)$, is determined by the Beer-Lambert law:
\begin{equation}
\tau_{\text {gas }}(\lambda) =\mathrm{e}^{-\alpha_{\text {gas }}(\lambda) \int_0^l \mathrm{c}(\mathrm{x}) \mathrm{dx}}
\end{equation}

In the equation, $\alpha_{gas}(\lambda)$ represents the spectral absorption coefficient of the gas, while $c(x)$ denotes the gas concentration (ppm) at the position x along the gas cloud path. The parameter $l$ signifies the total length (in meters) of the gas cloud path along the line of sight.

Based on this observation, it is indicated that higher gas concentrations in the real world lead to reduced gas spectral transmittance, resulting in a more distinct contrast in radiance between the gas and non-gas paths. As a consequence, the gas captured by the gas cloud imaging system stands out more prominently in comparison to the background. The aforementioned analysis demonstrates that changes in the pixel intensity can reflect the variation trend of gas concentration in the real world. Hence, in the feature representation phase, It is reasonable to approximate the Gaussian gas dispersion model through the voxel shift field.

\section{GOD-Video dataset Overview}

\begin{figure*}
	\vspace{-0.5em}
	\begin{center}
		\includegraphics[width=0.85\linewidth]{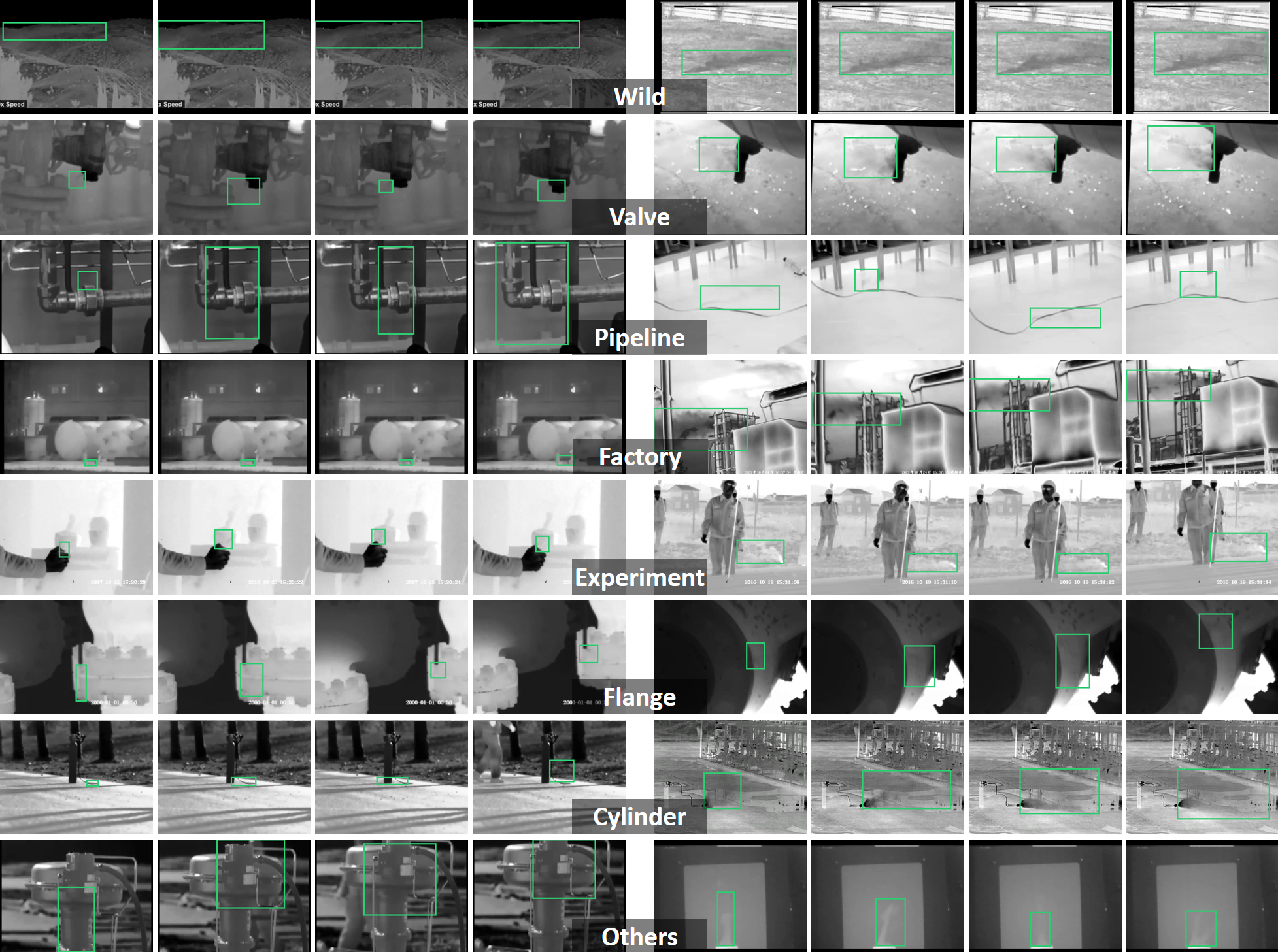}
	\end{center}
	\vspace{-1.5em}
	\caption{ The overview of the GOD-Video Dataset, including scenes: wild, valve, pipeline, factory, experiment, flange, cylinder and others. 
 }
	\label{fig:GOD-Video_vis}
		\vspace{-0.5em}
\end{figure*}

Fig.~\ref{fig:GOD-Video_vis} provides an overview of the GOD-Video dataset. It comprises scenes featuring active gas leak experiments, as exemplified by the experiment and cylinder sets. Additionally, scenes captured in industrial environments are classified into pipeline, flange, valve, and factory sets based on their respective backgrounds. Moreover, the wild set represents natural scenes, while the remaining scenes are categorized as others. On the left side, vague samples with challenging object boundary judgments are presented, whereas the right side displays clear samples with relatively well-defined boundaries.

\section{Optical Flow and Background Modeling Results}
\begin{figure*}
	\begin{center}
		\includegraphics[width=0.85\linewidth]{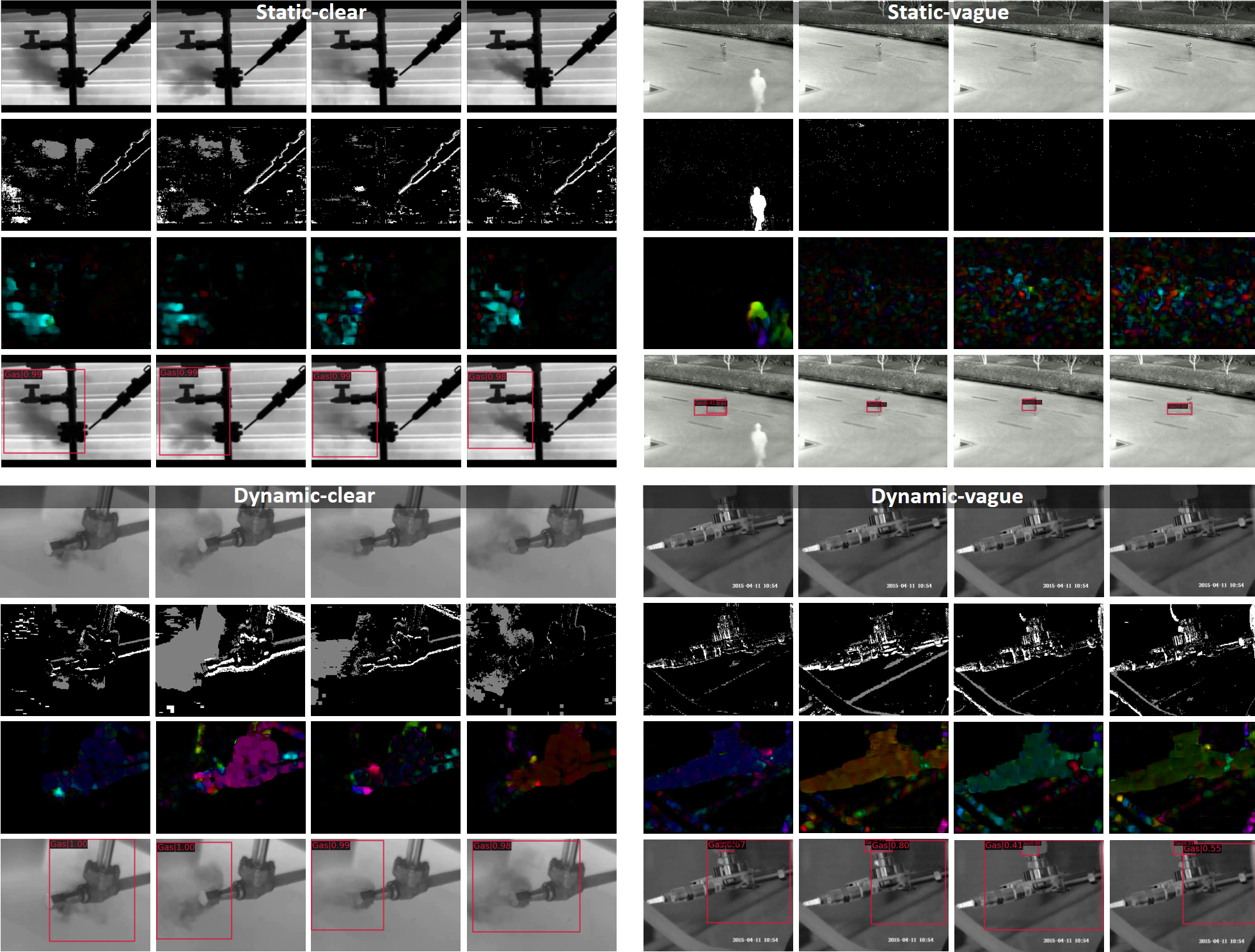}
	\end{center}
	\vspace{-1.5em}
	\caption{Each subgraph's first row displays the original video sequence, followed by the background modeling results in the second row, the dense flow results in the third row, and the detection results of VSF RCNN in the fourth row.
 }
	\label{fig:flow}
\end{figure*}

To explain the reasons behind the degradation of detector performance caused by the introduction of optical flow, we present visualizations of optical flow and background modeling algorithms. Fig.~\ref{fig:flow} reveals that optical flow and background modeling algorithms can approximately extract the leaked gas area solely under the premise of a stationary camera and extremely high gas concentration. Nevertheless, under the low gas concentration, the temporal and spatial variations of pixel values diminish significantly, posing challenges to capturing the spatio-temporal morphological changes of the gas. Additionally, during camera motion, optical flow reflects information from prominent objects in the background, rendering it incapable of extracting temporal variation. In contrast, our proposed VSF RCNN effectively overcomes the limitations of traditional methods, enabling detection of extremely weak gaseous objects in various complex environments.

\section{Objectness Measures}
\begin{figure*}
	\begin{center}
		\includegraphics[width=1.0\linewidth]{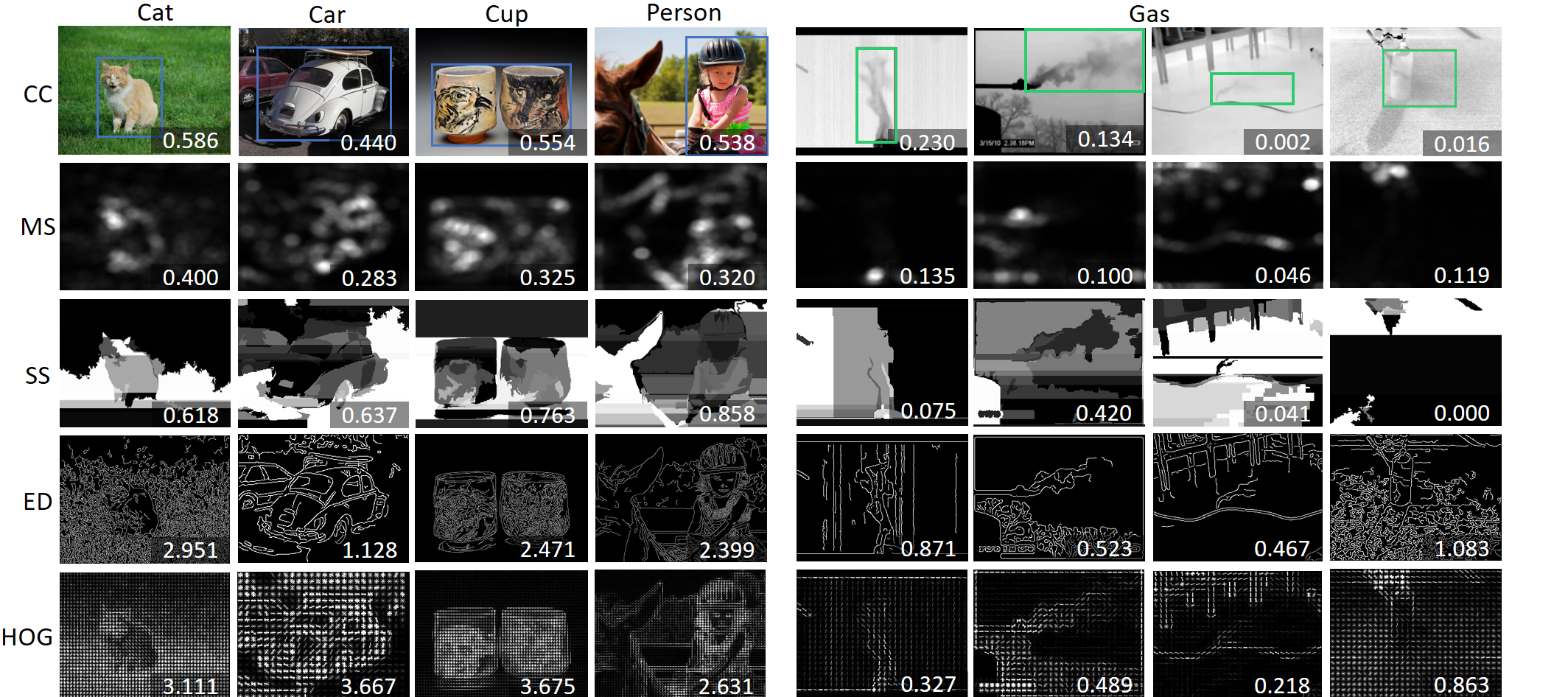}
	\end{center}
	\vspace{-1.5em}
	\caption{  More visualization results and scores of objectness measures between the conventional objects (cat, car, cup, person) and gaseous objects. 
 }
	\label{fig:obejctness}
		\vspace{-0.5em}
\end{figure*}

Fig.~\ref{fig:obejctness} presents additional results of objectness measures. The visualization results show the effectiveness of Superpixels Straddling (SS) and Edge Density (ED) in precisely delineating the boundaries of conventional objects, such as cats, cars, cups and people. These objects exhibit pronounced saliency and significant disparities between the object foreground and background. In contrast, for gaseous objects, regardless of sample clarity, SS and ED primarily capture background edges, posing challenges in delineating the boundaries of the gaseous object. The scores for MS and CC are also markedly low, and gaseous objects have arbitrary degree of freedom in shape. As a result, frame-level detectors presents considerable difficulties in the GOD task, requiring the integration of temporal information to compensate for spatial information deficiency.

\section{Visualization Comparisons of Detectors}
\begin{figure*}
	\begin{center}
		\includegraphics[width=0.90\linewidth]{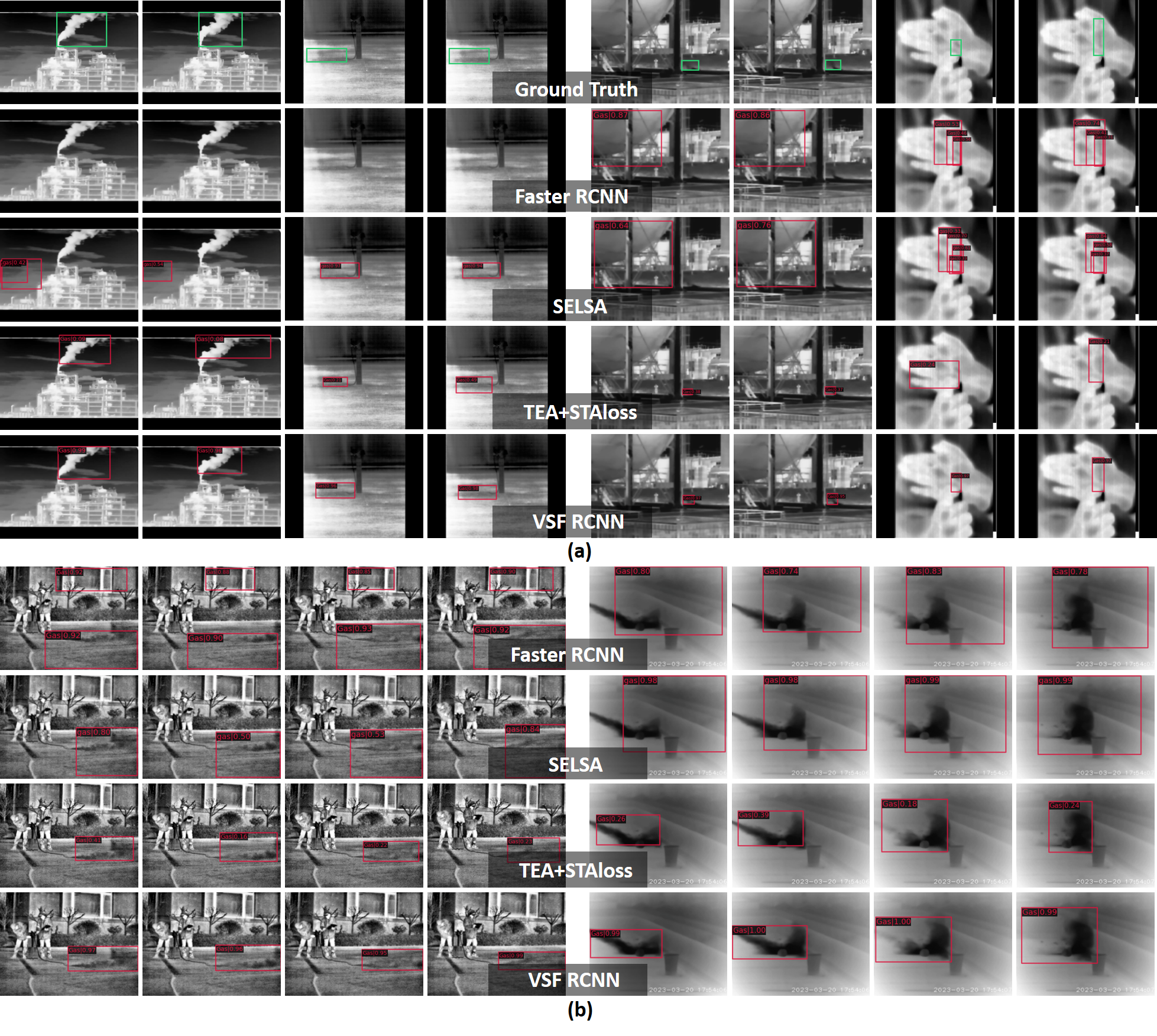}
	\end{center}
	\vspace{-1.5em}
	\caption{ (a) The comparison results of different detectors on the GOD-Video dataset. (b) The comparison results of different detectors on the test samples which are beyond the GOD-Video dataset.
 }
	\label{fig:sp_vsf_comp}
\end{figure*}

Here we present visualization comparisons of the Faster RCNN, SELSA, CenterNet (TEA+STAloss) and VSF RCNN in Fig.~\ref{fig:sp_vsf_comp}. Due to insufficient utilization of temporal information, Fast RCNN and SELSA lack the extraction of discriminative features, leading to false negatives, false positives and duplicate detection boxes. CenterNet (TEA+STAloss) and VSF RCNN exhibit satisfactory detection results, while VSF RCNN demonstrates superior localization accuracy.

Additionally, we test leakage samples beyond the GOD-Video dataset, and the utilization of low-cost, un-cooled infrared gas imaging system results in substantial image noise interference, impacting the performance of Faster RCNN and SELSA. CenterNet (TEA+STAloss) and VSF RCNN demonstrate the robustness and generalization ability to the low signal-to-noise ratio. Particularly, VSF RCNN excels in accurately delineating the boundaries of gaseous objects, highlighting its practical application value.

\vfill

\end{document}